%% file: main.tex
\documentclass[twocolumn,journal]{IEEEtran}
\usepackage[square, comma, sort&compress, numbers, sectionbib]{natbib}
\usepackage[sectionbib]{chapterbib}

\usepackage{arydshln}
\usepackage{multicol}
\usepackage{epsfig}
\usepackage{epstopdf}
\usepackage{graphicx}
\usepackage{citesort}
\usepackage{amssymb}
\usepackage{amsmath}
\usepackage{color}
\usepackage{url}
\usepackage{array}
\usepackage{multirow}
\usepackage[table]{xcolor}
\usepackage{bm}
\usepackage{tablefootnote}
\usepackage{threeparttable}
\usepackage{color}
\usepackage{colortbl}
\usepackage{mdwmath,mdwtab,amsmath}
\usepackage{autobreak}
\usepackage{amsmath,amssymb}
\usepackage{array}
\usepackage{multirow}
\usepackage{tabularx}
\usepackage{makecell}
\usepackage{graphicx}
\usepackage{threeparttable}
\usepackage{tablefootnote}
\usepackage[OT1]{fontenc}
\usepackage{float}
\newcolumntype{Y}{>{\centering\arraybackslash}X}
\usepackage[ruled,linesnumbered]{algorithm2e}
\usepackage{tabularx}
\usepackage[hang,flushmargin]{footmisc}
\usepackage{ragged2e}
\usepackage{fancyhdr}
\usepackage{xcolor}
\usepackage{bbding} 
\usepackage{indentfirst}
\usepackage{stfloats}
\usepackage{xpatch}
\usepackage{etoolbox}
\usepackage[noend]{algpseudocode}
\makeatletter
\usepackage[colorlinks, citecolor=green]{hyperref}
\usepackage{colortbl}  
\usepackage{xcolor}
\usepackage{array}   

\newlength{\figurewidth}
\newlength{\smallfigurewidth}

\usepackage{tikz,xcolor,hyperref}
\definecolor{lime}{HTML}{A6CE39}
\DeclareRobustCommand{\orcidicon}{
	\begin{tikzpicture}
		\draw[lime, fill=lime] (0,0)
		circle[radius=0.16]
		node[white]{{\fontfamily{qag}\selectfont \tiny \.{I}D}}; 
	\end{tikzpicture}
	\hspace{-2mm}
}
\foreach \x in {A, ..., Z}{%
	\expandafter\xdef\csname orcid\x\endcsname{\noexpand\href{https://orcid.org/\csname orcidauthor\x\endcsname}{\noexpand\orcidicon}}
}

\hypersetup{hidelinks}

\begin{document}

	\title
	{Task-wise Sampling Convolutions for Arbitrary Oriented Object Detection in Aerial Images}
	\author{%
		\vspace{0.5em}
		Zhanchao Huang,~\IEEEmembership{Member,~IEEE}, 
		Wei Li,~\IEEEmembership{Senior Member,~IEEE},
		Xiang-Gen Xia,~\IEEEmembership{Fellow,~IEEE},\\
		Hao Wang, 
		and
		Ran Tao,~\IEEEmembership{Senior Member,~IEEE}
		\thanks{%
			This work was supported by National Key R\&D Program of China under Grant 2021YFB3900502. (Corresponding Author: Wei Li; e-mail: liwei089@ieee.org)}
		\thanks{%
			Zhanchao Huang is with the Key Laboratory of Spatial Data Mining and Information Sharing of Ministry of Education, The Academy of Digital China, Fuzhou University, Fuzhou 350108, China; with the National \& Local Joint Engineering Research Center of Satellite Geospatial Information Technology, Fuzhou University, Fuzhou 350108, China; and with the School of Information and Electronics, Beijing Institute of Technology, and Beijing Key Lab of Fractional Signals and Systems, Beijing 100081, China. (e-mail: zhanchao.h@outlook.com)}
		\thanks{%
			Wei Li, Hao Wang, Ran Tao are with the School of Information and Electronics, Beijing Institute of Technology, and Beijing Key Lab of Fractional Signals and Systems, Beijing 100081, China. (e-mail: liwei089@ieee.org; haohaolalahao@icloud.com; rantao@bit.edu.cn)}
		\thanks{%
			Xiang-Gen Xia is with the Department of Electrical and Computer Engineering, University of Delaware, Newark, DE 19716, USA. (e-mail: xxia@ee.udel.edu)}
	}
\maketitle
\thispagestyle{fancy}
\renewcommand{\headrulewidth}{0pt} 
\pagestyle{fancy}

\begin{abstract}
\input{Abstract}
\end{abstract}


\section{Introduction}
\input{Chap1}

\section{Related Works}
\input{Chap2}

\section{Proposed GGHL Framework}
\input{Chap3}

\section{Experiments and Discussions}
\input{Chap4}

\section{Conclusions}
\input{Chap5}


\footnotesize
\bibliographystyle{IEEEtran}
\bibliography{TS-Conv.bib}

\newpage
\normalsize
\section*{Supplementary Materials}
\input{SuppMate}


\end{document}

%% file: Abstract.tex
Arbitrary-oriented object detection (AOOD) has been widely applied to locate and classify objects with diverse orientations in remote sensing images. However, the inconsistent features for the localization and classification tasks in AOOD models may lead to ambiguity and low-quality object predictions, which constrains the detection performance. In this paper, an AOOD method called task-wise sampling convolutions (TS-Conv) is proposed. TS-Conv adaptively samples task-wise features from respective sensitive regions and maps these features together in alignment to guide a dynamic label assignment for better predictions. Specifically, sampling positions of the localization convolution in TS-Conv is supervised by the oriented bounding box (OBB) prediction associated with spatial coordinates. While sampling positions and convolutional kernel of the classification convolution are designed to be adaptively adjusted according to different orientations for improving the orientation robustness of features. Furthermore, a dynamic task-aware label assignment (DTLA) strategy is developed to select optimal candidate positions and assign labels dynamicly according to ranked task-aware scores obtained from TS-Conv. Extensive experiments on several public datasets covering multiple scenes, multimodal images, and multiple categories of objects demonstrate the effectiveness, scalability and superior performance of the proposed TS-Conv.

%% file: Chap1.tex
\IEEEPARstart{A}{rbitrary-oriented} object detection (AOOD) is widely applied in remote sensing scenes \cite{xiaDOTALargeScaleDataset2018}. Distinguished from the objects in ordinary object detection (OD) \cite{renFasterRCNNRealTime2017a,bai2023localizing}, objects in AOOD have more diverse orientations and are localized by oriented bounding boxes (OBBs). Recently, benefiting from the intensive research on convolutional neural networks (CNNs), many CNN-based AOOD methods have emerged \cite{ding2019learning,xu2020gliding,yang2019r3det,huang2022general}.
\begin{figure}[tb]
	\vspace{-1em}
	\centering
	\epsfig{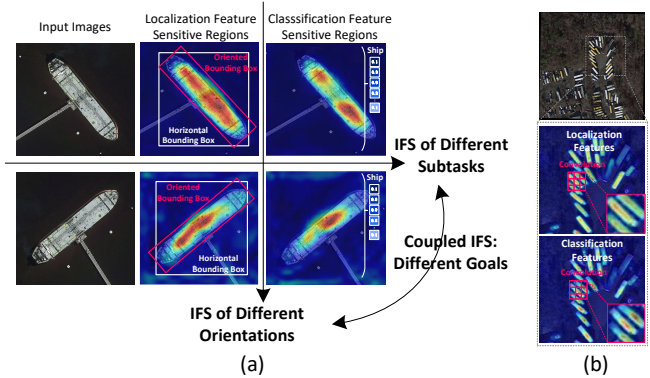}
	\caption{(a) The IFS problems of AOOD in different subtasks and orientations. (b) The IFS problem is exacerbated by the diverse orientation and dense distribution of objects. }\label{fig:2}
	\vspace{-1em}
\end{figure}

However, when locating and classifying objects in a CNN model, these two tasks may focus on different feature regions, as shown in Fig.~\ref{fig:2} (a). This problem of inconsistent feature sensitivity (IFS) between localization and classification tasks \cite{jiang2018acquisition,song2020revisiting,piao2023unsupervised} constrains the performance of object detection. Moreover, compared with the IFS problem in general object detection (GOD), the IFS problem in AOOD is more challenging in the sense that not only IFS of different subtasks but also IFS of different orientations should be considered, as well as the problem of coupling the two problems. On one hand, Compared with GOD, IFS of different orientations in AOOD should be considered. As shown in Fig.~\ref{fig:2} (a), when the orientations of the ships are different, their sensitive regions of localization features are also different. Since the bounding box in GOD does not consider the orientation, these ships can even be localized using similar bounding boxes predicted from different sensitive features \cite{huang2022general}. While in AOOD, the bounding box also has orientation characteristics, so IFS of different orientations has a great impact on the accuracy of object localization. On the other hand, AOOD needs to further consider the inconsistency in learning goals when IFS of different subtasks and IFS of different orientations problems are coupled. Different from the features and predictions of the localization subtask in AOOD that need to change with IFS of different orientations, the features of the classification subtask should be robust to IFS of different orientations without changing for the same category of objects. 
	
Therefore, when the IFS of different subtasks is coupled with IFS of different orientations, the learning goals of different subtasks for IFS are different. Furthermore, objects usually have more diverse orientations and dense distribution in AOOD tasks \cite{han2021redet}, which also exacerbates the IFS problem. Since objects of the same category have different orientations in the remote sensing overhead perspective, their IFS is not only between different subtasks and categories but also within the same category and subtasks. Besides, since many objects in remote sensing images may be densely distributed \cite{yangSCRDetMoreRobust2019}, the feature-sensitive areas of different objects may interfere with the feature extraction of vanilla convolution. As shown in Fig.~\ref{fig:2} (b), since the vanilla convolution neighborhood range is fixed, surrounding objects may interfere with feature extraction within the convolution area, i.e., the difference between the two feature-sensitive regions in the two red boxes in Fig.~\ref{fig:2} (b) becomes larger. This exacerbates IFS, which is not directly due to subtask or orientation differences. In addition, the IFS problem also leads to ambiguity in label assignment due to the different preferences of each candidate for location-sensitivity or classification-sensitivity.

In response to the IFS problem of different subtasks, RepPoints \cite{yang2019reppoints}, Oriented RepPoints \cite{li2022oriented}, etc., sampled localization and classification features by two DCNs that share the aligned sampling offsets in two CNN branches. However, the positions suitable for sampling localization features may not be suitable for sampling classification features. The forced alignment of sampling offsets for different subtasks may cause some sampling points that are only suitable for one task to be missed. Furthermore, a candidate position for accurate localization may not be optimal for object classification when CNN makes predictions. In this regard, AutoAssign\cite{zhu2020autoassign}, ATSS\cite{zhang2020bridging}, GGHL \cite{huang2022general}, etc., indirectly improve the prediction by adjusting the selection or weights of positive and negative samples to optimize the training supervision. However, these methods do not directly change the feature extraction or alignment process, the IFS problem existed in the sampling process of the convolutional layer is not eliminated. However, AOOD not only faces IFS of different subtasks but also faces IFS of different directions, as well as the problem of coupling the two IFS together. Existing researches do not consider these and other challenges.

Rethinking the above problems and observing Fig.~\ref{fig:2}, a straightforward idea is 
to separately extract more appropriate features from different sampling positions that are suitable for different subtasks and different object orientations by different sampling strategies. Following this idea, two more aspects have to be carefully considered: 1) selecting appropriate sampling strategies for extracting task-wise features, and 2) designing association constraints for different sampling offsets since they are not entirely independent. Besides, in the design of the task-wise convolutional sampling scheme, the characteristics of objects, including OBB representation, dense distribution, diverse orientation, etc., need to be carefully considered. Furthermore, when the task-wise features are mapped to the same candidate position for predictions, it requires a dynamic label assignment strategy to provide feedback to the CNN on which candidate positions are suitable. Therefore, motivated by the above considerations of the IFS problem and AOOD challenges, a Task-wise Sampling Convolutions (TS-Conv) method is proposed in this work. The main contributions are summarized as follows.

1) The IFS problem of different subtasks and different orientations in AOOD is comprehensively analyzed. On this basis, the proposed TS-Conv designs more explicitly supervised sampling strategies to extract the task-wise features from appropriate sensitive regions. Furthermore, the sampling strategies, feature alignment processes, and label assignment are unified into a closely related framework to achieve dynamic feedback optimization.

2) In the designed TS-Conv, the localization sampling offsets are directly associated with the spatial coordinate embeddings and the OBB decoding in the detection head, which change with the orientations and sizes of objects. The classification sampling offsets are rotation invariant and can be used with the designed dynamic circular kernel (DCK) to extract orientation-robust features by adjusting the optimal orientation and weights.

3) Based on the predictions of TS-Conv, a dynamic task-consistent-aware label assignment (DTLA) strategy is developed, in which the aligned features are adaptively mapped to positive samples selected by the ranked task-consistent-aware score. It forms a dynamic closed loop of "assignment-sampling-alignment-reassignment" for supervising CNN training to obtain better performance.

The remainder of this paper is organized as follows. Section II reviews the related works. Section III presents the proposed TS-Conv in detail. Section IV evaluates and analyzes the performance of TS-Conv through extensive experiments. Section V draws the conclusions and discussions.

%% file: Chap2.tex
\subsection{Arbitrary-Oriented Object Detection}
Unlike ordinary OD, the objects in AOOD are positioned by OBBs. Thus, additional CNN localization branches with different OBB representation strategies are designed to predict OBBs. ROI Transform \cite{ding2019learning}, SCR-Det \cite{yangSCRDetMoreRobust2019}, R3Det \cite{yang2019r3det}, S$^2$ANet \cite{han2021align}, etc., predicted rotation angles based on horizontal bounding boxes (HBBs) to obtain OBBs, while Gliding Vertex \cite{xu2020gliding}, GGHL \cite{huang2022general}, etc. directly predicted the four-vertices of each OBB. Oriented RepPoints \cite{li2022oriented} represented OBBs by a set of points. Distinguishing from the above angle or vertex regression strategies, CSL \cite{yang2020arbitrary} predicted the rotation angles of OBBs by discrete angle classification. To get rid of the dependence on anchor boxes, BBAVectors \cite{yi2020oriented}, O$^2$-DNet \cite{wei2020oriented}, GGHL \cite{huang2022general}, etc., developed different anchor-free label assignment strategies, and AO2-DETR \cite{dai2022ao2} predicted OBBs by the sequence model without anchor boxes. For more accurate OBB predictions, GWD \cite{yang2021rethinking} and KLD \cite{yang2021learning} designed new loss functions based on the distances of Gaussian distributions generated from the predicted OBBs and ground truth. CFA \cite{9578090} proposed a convex-hull feature adaptation method, which optimizes feature assignment by constructing convex-hull sets and dynamically splitting positive and negative convex-hulls. DCFL \cite{Xu_2023_CVPR} proposed a dynamic prior along with the coarse-to-fine assigner to leverage the coarse prior matching and finer posterior constraint to dynamically assign labels, which provides appropriate and relatively balanced supervision for diverse instances. In response to the IFS problem in AOOD, GGHL \cite{huang2022general}, CFC-Net \cite{ming2021cfc}, etc., decoupled the CNN branches and reweighted task-wise features by different strategies. Oriented RepPoints \cite{li2022oriented} predicted initial OBBs and categories through aligned features extracted by shared-offsets DCNs and then learned additional offsets to refine the initial OBBs. Different from the existing AOOD methods, the proposed TS-Conv separately samples task-wise features without additional reweighting or refinement operation. Besides, TS-Conv associates the task-wise feature sampling with other parts of the AOOD pipeline, including the OBB representation and label assignment.

\subsection{Solutions for the IFS Problem}
IoU-Net \cite{jiang2018acquisition} analyzed the IFS problem for the first time, which predicted an additional localization score and aggregated it with the classification score as the final score. Along with this idea of post-processing at the prediction side, IoU-aware \cite{zhang2021varifocalnet}, GGHL \cite{huang2022general}, etc., designed different strategies to obtain prediction scores that consider both localization and classification contributions by re-weighting, re-ranking, or jointly optimizing the scores, etc. Nevertheless, the feature-sensitive positions are still spatial misalignment. Double-Head R-CNN \cite{wu2020rethinking}, YOLOX \cite{ge2021yolox}, etc., decoupled the detection head of CNN into two branches to extract localization and classification features separately. Based on this feature-decoupling scheme, two-stage methods, TSD \cite{song2020revisiting}, D2Det \cite{cao2020d2det}, etc., extracted task-suitable features by task-wise Deformable RoI Pooling \cite{zhu2019deformable}. One-stage methods, such as Guided Anchoring \cite{wang2019region}, VFNet \cite{zhang2021varifocalnet}, etc., aligned and refined the decoupled features by DCNs \cite{dai2017deformable,zhu2019deformable} or attention mechanism. Furthermore, RepPoints \cite{yang2019reppoints} and Oriented RepPoints \cite{li2022oriented} extracted localization and classification features from the spatially aligned positions in two branches by two DCNs that share the same sampling offsets. In addition, Oriented RepPoints \cite{li2022oriented} designed a quality assessment and sample assignment strategy for selecting high-quality sampling points and positions. However, the sampling positions of localization and classification features in the above methods are spatially aligned. Inspired by the above contradiction, this work designs separately sampling strategies to solve the IFS problem in AOOD. Moreover, TS-Conv also associates OBB representation and dynamic label assignment with task-wise feature sampling, which deals with the IFS problem more comprehensively from different perspectives of the AOOD framework.

%% file: Chap3.tex
\begin{figure}[tp]
	\centering
	\epsfig{width=0.48\textwidth,file=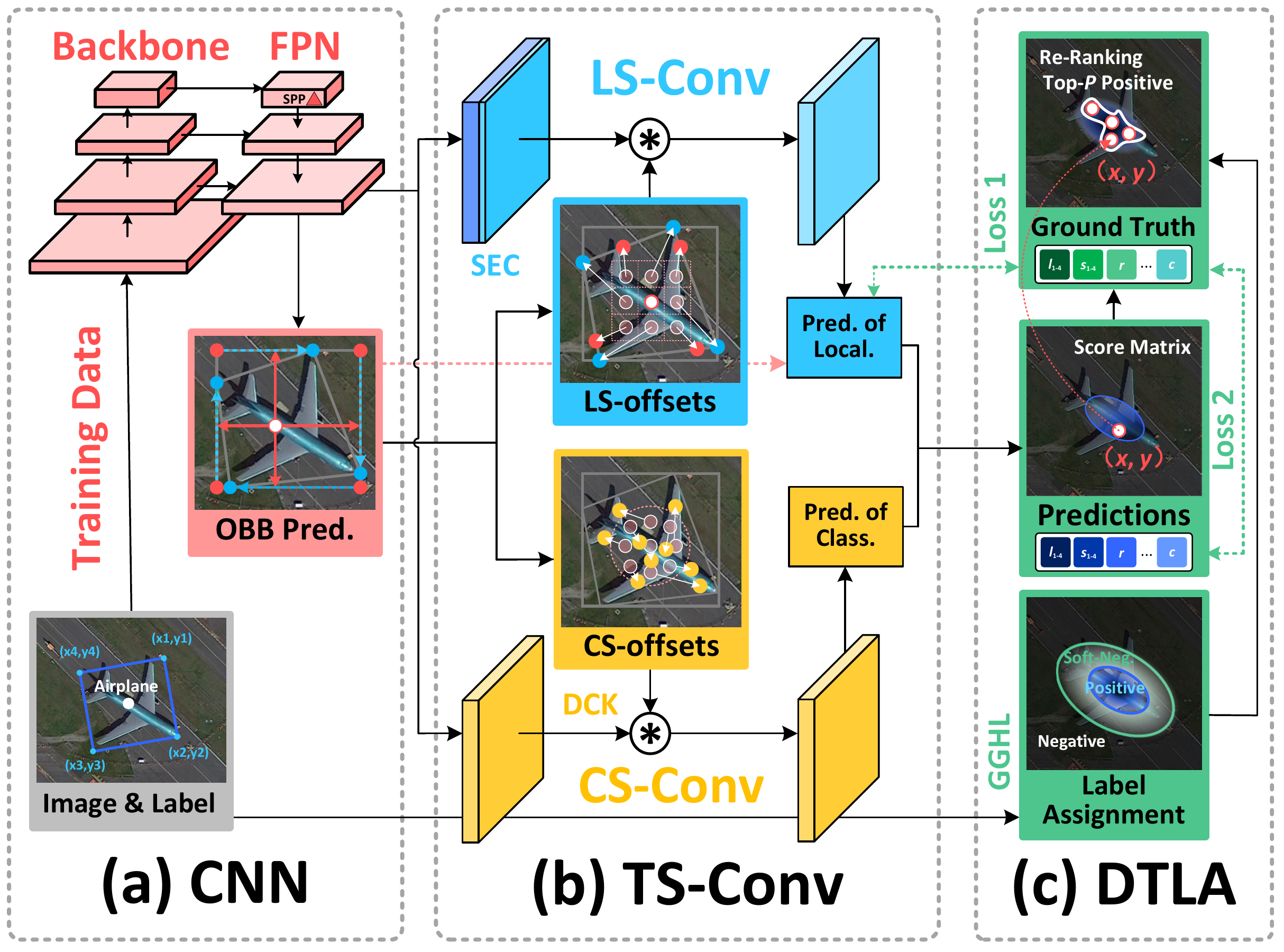}
	\caption{The TS-Conv framework comprises (a) the CNN model and training data, (b) the proposed TS-Conv consisting of LS-Conv and CS-Conv, (c) the designed DTLA strategy. The baseline CNN model used by TS-Conv and GGHL is Darknet53+FPN. GGHL in the following (when the label assignment strategy is not emphasized) refers to the GGHL-based Darknet53+FPN model.}\label{fig:3}
\end{figure}

The AOOD framework based on TS-Conv is shown in Fig.~\ref{fig:3}, which is mainly composed of three parts: (a) the CNN model and training data, (b) the proposed TS-Conv component consisting of the convolution for sampling localization features (named as LS-Conv) and the convolution for sampling classification features (named as CS-Conv), (c) the designed DTLA strategy.

First, the CNN extracts features through the backbone and FPN and outputs them to the dual branches of localization and classification in the TS-Conv component, marked in blue and yellow in Fig.~\ref{fig:3} (b). It follows the scheme of decoupling branches in Fig.~\ref{fig:2} (b). Meanwhile, initial OBBs, marked in red in Fig.~\ref{fig:3} (a), are predicted as the constraint to supervise the learning of sampling positions of TS-Conv. In response to the problems of the shared-offsets DCNs scheme in Fig.~\ref{fig:2} (c), different sampling offsets constrained by initial OBBs are designed for localization and classification, respectively. Then, the convolutions for localization and classification use the designed sampling strategies to extract features from their sensitive regions separately. It allows the task-wise sampling positions of CNN to be supervised by ground truth to adapt to different objects dynamically. Next, aiming at the problem that static label assignment strategies cannot give CNN feedback on assigning candidate positions without the IFS problem for predictions during training, a dynamic label assignment strategy DTLA is designed. In DTLA, the static arbitrary-oriented label assignment strategy GGHL \cite{huang2022general} is utlized as the initial assignment of DTLA. Based on the initial assignment, the designed dynamic label assignment strategy, i.e., DTLA, is carried out. As shown in the green in Fig.~\ref{fig:3} (c), DTLA adjusts the positive and negative samples pre-assigned by the static label assignment method GGHL \cite{huang2022general} according to the task-consistent-aware score predicted in each iteration of TS-Conv training. Finally, the CNN is trained to make the loss between CNN’s predictions and labels assigned by DTLA converge, and then the trained CNN is used to detect objects.

\subsection{Convolution for Sampling Localization Features}
\begin{figure}[tbp]
	\centering
	\epsfig{width=0.4\textwidth,file=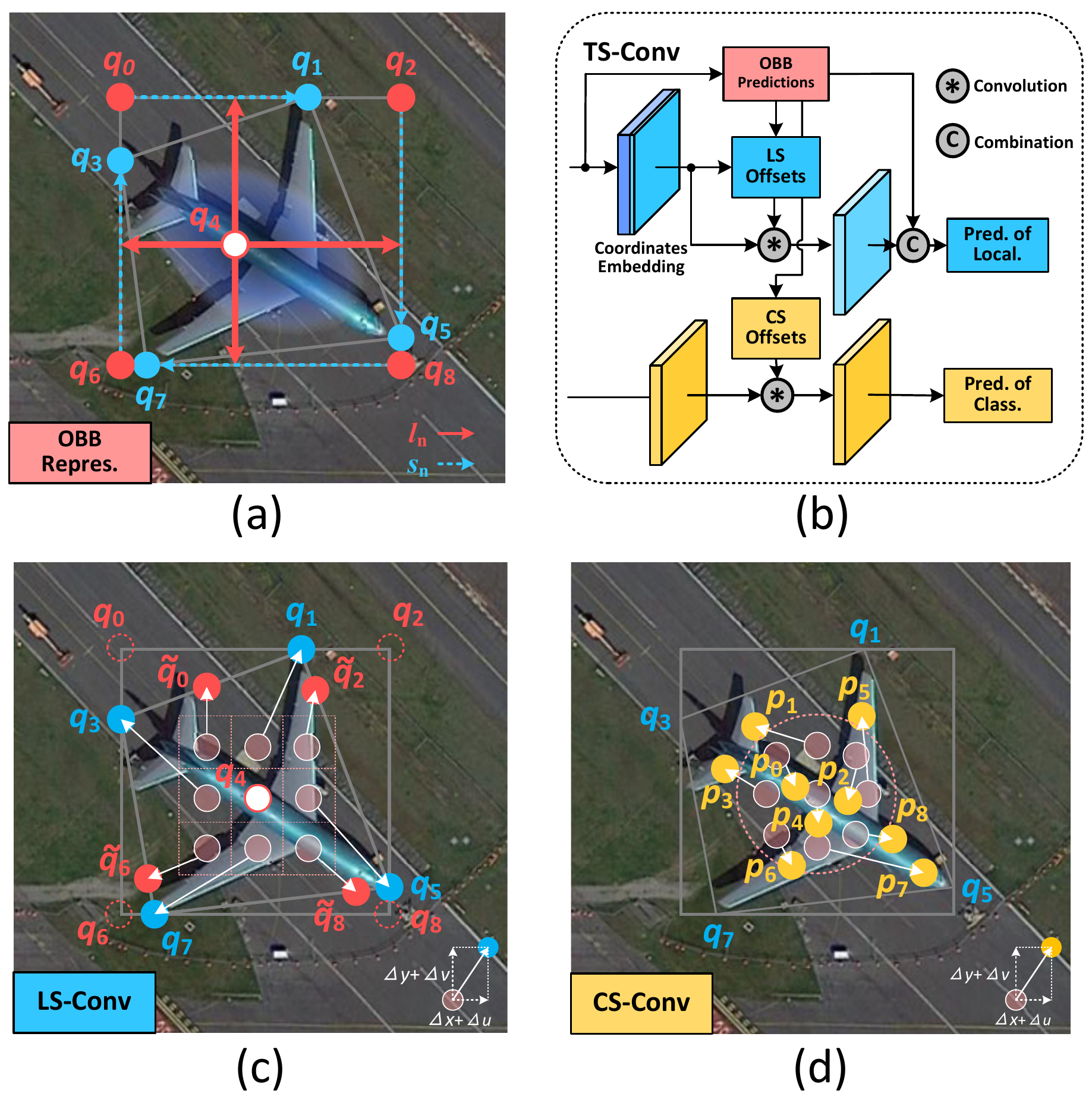}
	\caption{The principle of the proposed task-wise sampling convolutions (TS-Conv). (a) The OBB representation of GGHL \cite{huang2022general}. (b) The CNN structure of the proposed TS-Conv. (c) The sampling positions of the convolution for sampling localization features (LS-Conv). (d) The sampling positions of the convolution for sampling classification features (CS-Conv).}
	\label{fig:4}
	\vspace{-0.5em}
\end{figure}

From the observation in Fig.~\ref{fig:2} (a) and the analysis of \cite{song2020revisiting}, the feature-sensitive regions for localization are ainly distributed at the object boundary, i.e., the regions near the OBB. Inspired by this observation and the star-shaped box of VFNet \cite{zhang2021varifocalnet}, associating the feature sampling positions of the LS-Conv with the OBB representation is considered. For OBB representation, GGHL \cite{huang2022general} proposed a flexible anchor-free strategy based on Gliding Vertex \cite{xu2020gliding}, as shown in Fig.~\ref{fig:4} (a). It predicts an OBB by predicting the distances ${l_n},{\rm{ }}n = 1,2,3,4,$ from the current candidate object position $q_4\left( {x,y} \right)$ to the four edges of the external horizontal bounding box (HBB) and the distances ${s_n},{\rm{ }}n = 1,2,3,4,$ from the four vertices $q_0$, $q_2$, $q_6$, and $q_8$ of the HBB to the four vertices $q_1$, $q_3$, $q_5$, and $q_7$ of the OBB on the corresponding edge. This OBB representation has nine key positions, i.e., the four vertices of the HBB, the four vertices of the OBB, and the current candidate object position. A straightforward idea is to directly correspond these key positions to the nine sampling points of a $3 \times 3$ convolution, as shown in Fig.~\ref{fig:4} (c). The localization convolution no longer explores the features in the current position and its eight neighborhoods fixedly while adjusting the sampling position according to different objects with different OBBs by DCN \cite{pan2020dynamic} dynamically. The details and further improvements are presented below.

First, the initial OBBs are predicted by the CNN model trained under the supervision of labels assigned by GGHL \cite{huang2022general}. The initial estimations of ${l_n}$ and ${s_n,{\rm{ }}n = 1,2,3,4,}$ are denoted as ${\hat l_n}$ and ${\hat s_n,{\rm{ }}n = 1,2,3,4}$, respectively. In this case, ${\hat s_1},{\hat s_3} \in \left[ {0,{{\hat l}_2} + {{\hat l}_4}} \right]$ and ${\hat s_2},{\hat s_4} \in \left[ {0,{{\hat l}_1} + {{\hat l}_3}} \right]$. Second, the spatial coordinates of the sampling points ${q_i},$ $i = 0,1, \cdots ,8,$ are calculated according to the predictions ${\hat l_n}$, ${\hat s_n}$, and the coordinates of the current convolutional position $\left( {x,y} \right)$. In this case shown in Fig.~\ref{fig:4} (a), the current sample point is defined as ${q_4}=\left( x, y \right)$. The four vertices of HBB are defined as ${q_0}=\left( {x - {{\hat l}_4},} \right.$ $\left. {y - {{\hat l}_1}} \right)$, ${q_2}=\left( {x + {{\hat l}_2},} \right.$ $\left. {y - {{\hat l}_1}} \right)$, ${q_6}=\left( {x - {{\hat l}_4},} \right.$ $\left. {y + {{\hat l}_3}} \right)$ and ${q_8}=\left( {x + {{\hat l}_2},} \right.$ $\left. {y + {{\hat l}_3}} \right)$, respectively. The four vertices of OBB are defined as ${q_1}=\left( {x - {{\hat l}_4} + {s_1},}\right.$ $\left. {y - {{\hat l}_1}} \right)$, ${q_3}=\left( {x - {{\hat l}_4},} \right.$ $\left. {y + {{\hat l}_3} - {{\hat s}_4}} \right)$, ${q_5}=\left( {x + {{\hat l}_2},} \right.$ $\left. {y - {{\hat l}_1} + {{\hat s}_2}} \right)$ and ${q_7}=\left( {x + {{\hat l}_2} - {{\hat s}_3},} \right.$ $\left. {y + {{\hat l}_3}} \right)$, respectively. Intuitively, the process of manually labeling the bounding box of an object by human is usually to find the most prominent points on the top, bottom, left, and right of an object, and then draw the bounding box based on this range. That is, for an ideal bounding box label that closely surrounds an object, at least one point on each edge intersects the object. Thus, the intersection points between the four sides of the bounding box and the object are the key elements for locating the position and range of an object. Since objects have various shapes and each edge may have more than one intersection point with the object, a sampling point on each of the four sides is selected and set to be adaptive sliding. The sliding processes are controlled by the CNN learnable variables ${\sigma _i} \in \left( {0,1} \right),{\rm{ }}i = 0,2,6,8$.. These four sliding points are defined as ${\tilde q_0},{\tilde q_2},{\tilde q_6},{\tilde q_8},$, the coordinates them are listed in Table~\ref{table:1}. For convenience, let ${\tilde q_i} = {q_i},{\rm{ }}i = 1,3,4,5,7,$ in the following.
\begin{table}[tbp]
	\centering
	\renewcommand\arraystretch{1}
	\caption{{Summary of sampling points ${\tilde q_i},{\rm{ }}i = 0,2,6,8,$ of the designed LS-Conv}}
	\label{table:1}
	\setlength{\tabcolsep}{2mm}{	
		\resizebox{0.48\textwidth}{!}{
			\begin{tabular}{m{1cm}<{\centering}|m{5cm}<{}|m{2.5cm}<{\centering}}
				\hline\hline
				Points & \centering Coordinates & Binding Conditions \\ 
				\hline
				\multirow{4}{1cm}{\centering ${\tilde q_0}$} & $\left\{ \begin{array}{l}
					{x_{{{\tilde q}_0}}} = {x_{{q_1}}}\\
					{y_{{{\tilde q}_0}}} = {y_{{q_1}}} + {\sigma _0} \times \left( {{y_{{q_3}}} - {y_{{q_1}}}} \right)
				\end{array} \right.$ & ${\rm{ }}{x_{{q_1}}} = {x_{{q_0}}}$ \\
				\cline{2,3}
				~ & $\left\{ \begin{array}{l}
					{x_{{{\tilde q}_0}}} = {x_{{q_0}}} + {\sigma _0} \times \left( {{x_{{q_1}}} - {x_{{q_0}}}} \right)\\
					{y_{{{\tilde q}_0}}} = {y_{{q_1}}} + \frac{{{y_{{q_3}}} - {y_{{q_1}}}}}{{{x_{{q_3}}} - {x_{{q_1}}}}} \times \left( {{x_{{{\tilde q}_0}}} - {x_{{q_1}}}} \right)
				\end{array} \right.$
				& ${x_{{q_0}}} < {x_{{q_1}}} \le {x_{{q_2}}}$ \\
				\hline
				\multirow{4}{1cm}{\centering ${\tilde q_2}$} & $\left\{ \begin{array}{l}
					{x_{{{\tilde q}_2}}} = {x_{{q_1}}}{\rm{    }}\\
					{y_{{{\tilde q}_2}}} = {y_{{q_2}}} + {\sigma _2} \times \left( {{y_{{q_5}}} - {y_{{q_2}}}} \right)
				\end{array} \right.$ & ${x_{{q_1}}} = {x_{{q_2}}}$ \\
				\cline{2,3}
				~ & $\left\{ \begin{array}{l}
					{x_{{{\tilde q}_2}}} = {x_{{q_2}}} - {\sigma _2} \times \left( {{x_{{q_2}}} - {x_{{q_1}}}} \right),{\rm{    }}\\
					{y_{{{\tilde q}_2}}} = {y_{{q_1}}} + \frac{{{y_{{q_5}}} - {y_{{q_1}}}}}{{{x_{{q_5}}} - {x_{{q_1}}}}} \times \left( {{x_{{{\tilde q}_2}}} - {x_{{q_1}}}} \right)
				\end{array} \right.$
				& ${x_{{q_0}}} \le {x_{{q_1}}} < {x_{{q_2}}}$ \\
				\hline			
				\multirow{4}{1cm}{\centering ${\tilde q_6}$} & $\left\{ \begin{array}{l}
					{x_{{{\tilde q}_6}}} = {x_{{q_7}}}\\
					{y_{{{\tilde q}_6}}} = {y_{{q_6}}} - {\sigma _6} \times \left( {{y_{{q_7}}} - {y_{{q_3}}}} \right)
				\end{array} \right.$ & ${x_{{q_6}}} = {x_{{q_7}}}$ \\
\cline{2,3}
~ & $\left\{ \begin{array}{l}
	{x_{{{\tilde q}_6}}} = {x_{{q_6}}} + {\sigma _6} \times \left( {{x_{{q_7}}} - {x_{{q_6}}}} \right)\\
	{y_{{{\tilde q}_6}}} = {y_{{q_7}}} + \frac{{{y_{{q_3}}} - {y_{{q_7}}}}}{{{x_{{q_3}}} - {x_{{q_7}}}}} \times \left( {{x_{{{\tilde q}_6}}} - {x_{{q_7}}}} \right)
\end{array} \right.$
& ${x_{{q_6}}} < {x_{{q_7}}} \le {x_{{q_8}}}$ \\
\hline
	\multirow{4}{1cm}{\centering ${\tilde q_8}$} & $\left\{ \begin{array}{l}
					{x_{{{\tilde q}_8}}} = {x_{{q_7}}}{\rm{  }}\\
					{y_{{{\tilde q}_8}}} = {y_{{q_7}}} - {\sigma _8} \times \left( {{y_{{q_7}}} - {y_{{q_5}}}} \right)
				\end{array} \right.$ & ${x_{{q_7}}} = {x_{{q_8}}}$ \\
\cline{2,3}
~ & $\left\{ \begin{array}{l}
	{x_{{{\tilde q}_8}}} = {x_{{q_8}}} - {\sigma _{{p_7}}} \times \left( {{x_{{q_8}}} - {x_{{q_7}}}} \right){\rm{   }}\\
	{y_{{{\tilde q}_8}}} = {y_{{q_7}}} + \frac{{{y_{{q_5}}} - {y_{{q_7}}}}}{{{x_{{q_5}}} - {x_{{q_7}}}}} \times \left( {{x_{{{\tilde q}_8}}} - {x_{{q_7}}}} \right)
\end{array} \right.$
& ${x_{{q_6}}} \le {x_{{q_7}}} < {x_{{q_8}}}$ \\
				\hline\hline
	\end{tabular}}}	
\end{table}

Define the feature map input to the localization branch as ${\boldsymbol{I}^{loc}} \in {\mathbb{R}^{W \times H \times F}}$, where $W,\ H,$ and $F$ denote the width, height, and the number of feature maps, respectively. Furthermore, inspired by extracting features from keypoint while letting CNN learn their position information, a spatial coordinate embedding operation is employed on LS-Conv. Generate a tensor ${\boldsymbol{I}^{coor-x}} \in {\mathbb{N}^{W \times H \times 1}}$. All the elements on each column of ${\boldsymbol{I}^{coor-x}}$ are the same, i.e., the index of this column. Similarly, generate a tensor ${\boldsymbol{I}^{coor-y}} \in {\mathbb{N}^{W \times H \times 1}}$. All the elements on each row of ${\boldsymbol{I}^{coor-y}}$ are the same, i.e., the index of this row. Define ${\tilde{\boldsymbol{I}}^{loc}} \in {\mathbb{R}^{W \times H \times (F+2)}}$ as the tensor of ${{\boldsymbol{I}}^{loc}}$, ${\boldsymbol{I}^{coor-x}}$, and ${\boldsymbol{I}^{coor-y}}$, i.e., ${\tilde{\boldsymbol{I}}^{loc}} \left({1:W}, {1:H}, {1:F}\right)\!={\boldsymbol{I}}^{loc}$, ${\tilde{\boldsymbol{I}}^{loc}} \left({1:W}, {1:H}, {F+1}\right)\!={\boldsymbol{I}}^{coor-x}$, ${\tilde{\boldsymbol{I}}^{loc}} \left({1:W}, {1:H}, {F+2}\right)\!={\boldsymbol{I}}^{coor-y}$. Define the element of ${\boldsymbol{\tilde I}^{loc}}$ at $\left( {x,y} \right)$, $x \in \left[ {1,W} \right]$ and $y \in \left[ {1,H} \right]$, in the first two dimensions  as $\boldsymbol{\tilde I}_{x,y}^{loc} \in \mathbb{R}^{1\times1\times(F+2)}$, which consists of a position and its associated $F$-dimensional feature vector. It allows the CNN to learn the spatial coordinates of the sampled positions directly while extracting the features from sensitive regions \cite{liu2018intriguing}. In addition, because the size $F$ of the feature vector satisfies $F \gg 2$ in CNN, the spatial coordinate embedding operation does not add much additional computational burden. Thus, the feature information and coordinates of sampling positions are explicitly and spatially corresponding to the feature map and its grid coordinates, and the geometric constraints between sampling positions are also clear and can be represented by the offsets of DCNs. Compared with using completely dynamic sampling positions, the learning goal of CNN is clearer.

For notational convenience, we define the correspondence of the nine elements (positions) of the following set 
\begin{equation}
\begin{array}{l}
	\left\{ {\left( { - 1, - 1} \right),\left( { - 1,0} \right)} \right.,\left( { - 1,1} \right),\left( {0, - 1} \right),\\
	\left( {0,0} \right),\left( {0,1} \right),\left. {\left( {1, - 1} \right),\left( {1,0} \right),\left( {1,1} \right)} \right\}\\
	\buildrel \Delta \over = \left\{ {\left( {\Delta x_u^{loc},\Delta y_u^{loc}} \right),u = 0,1, \cdots ,8} \right\}.
\end{array}
	\label{eq:0_1}
\end{equation}
The indices of $\left( {x,y} \right)$ and its eight neighbors are represented as $\left( {x + \Delta x_u^{loc},{\rm{ }}y + \Delta y_u^{loc}} \right)$, $u = 0,1, \cdots ,8$. Define the feature vectors at $\left( {x + \Delta {x_u^{loc}},{\rm{ }}y + \Delta {y_u^{loc}}} \right)$ as $\boldsymbol{\tilde I}_{x + \Delta {x_u^{loc}},y + \Delta {y_u^{loc}}}^{loc}$. Define a $3\times3$ filter $\boldsymbol{K}^{loc}$,
where $K_j^{loc},{\rm{ }}j = 0,1, \cdots ,8,$ are the elements and also the coefficients of $\boldsymbol{K}^{loc}$. When using $\boldsymbol{K}^{loc}$ to perform LS-Conv\footnote[1]{Note that, it is conventionally called a convolution operation in CNN, but it is actually a filtering operation.} on ${\boldsymbol{\tilde I}^{loc}}$, the following two cases are considered. When position $\left( {x,y} \right)$ is a positive position, which will be explained in Section III-C, used to predict the OBB, as shown in Fig.~\ref{fig:4} (c), the nine sampling points $\left( {x + \Delta {x_u^{loc}},{\rm{ }}y + \Delta {y_u^{loc}}} \right),{\rm{ }}u = 0,1, \cdots ,8,$ are correspondingly moved to the points $\left( {{x_{{{\tilde q}_i}}},{y_{{{\tilde q}_i}}}} \right),{\rm{ }}i = 0,1, \cdots ,8,$ by DCN \cite{dai2017deformable}. The sampled feature vectors are denoted as $\boldsymbol{\tilde I}_{{x_{{q_i}}},{y_{{q_i}}}}^{loc},{\rm{ }}i = 0,1, \cdots ,8$. When the position $\left( {x,y} \right)$ is not a positive position, the sampling points $\left( {x + \Delta {x_u^{loc}},{\rm{ }}y + \Delta {y_u^{loc}}} \right),{\rm{ }}u = 0,1, \cdots ,8,$ do not move. Thus, the output of LS-Conv at $\left( {x,y} \right)$ is represented as
\begin{equation}
\! \boldsymbol{O}_{x,y}^{loc} \! = \! \left\{ \begin{array}{l}
	\! \sum\limits_{\! i, j = 0}^{\! N = 8} { {\boldsymbol{\tilde I}_{{x_{{\tilde q_i}}},{y_{{\tilde q_i}}}}^{loc} K_j^{\! loc}} } m_j^{\! loc},{\enspace \rm{if}}\ (x,y){\ \rm{ is \ positive}},\\
	\! \sum\limits_{u, j = 0}^{N = 8} {\boldsymbol{\tilde I}_{x + \Delta {x_u^{\! loc}},y + \Delta {y_u^{\! loc}}}^{loc} K_j^{\! loc} m_j^{\! loc}} ,{\enspace \rm{otherwise,}}
\end{array} \right.
	\label{eq:2}
\end{equation}
where $\boldsymbol{O}_{x,y}^{loc} \in \mathbb{R}^{1\times1\times(F+2)}$ and CNN-learnable scalars $m_j^{loc} \in \left( {0,1} \right),{\rm{ }}j = 0,1, \cdots ,8,$ are employed to adjust the contribution of the features sampled from different positions like DCNv2 \cite{zhu2019deformable}. It further enhances the feature learning capability of LS-Conv by simultaneously adjusting the sampling position and amplitude. For some OBB vertices that do not fall on the object, it can suppress the sampling of low-quality features by decreasing the value of $m_j^{loc}$. Finally, the output features are used to predict the corrections $\Delta {\hat l_n}$ and $\Delta {\hat s_n},{\rm{ }}n = 1,2,3,4,$ to the initial OBB predictions ${\hat l_n}$ and ${\hat s_n},{\rm{ }}n = 1,2,3,4$. The refined OBB predictions are
\begin{equation}
	\left\{ \begin{array}{l}
		{{\tilde l}_n} = {{\hat l}_n} \times \Delta {{\hat l}_n}\\
		{{\tilde s}_n} = {{\hat s}_n} \times \Delta {{\hat s}_n}
	\end{array} \right.,
	\label{eq:3}
\end{equation}
where ${\tilde s_1},{\tilde s_3} \in \left[ {0,{{\tilde l}_2} + {{\tilde l}_4}} \right]$ and ${\tilde s_2},{\tilde s_4} \in \left[ {0,{{\tilde l}_1} + {{\tilde l}_3}} \right]$. Note that, due to the significant difference in the sizes of different objects, i.e., the values of ${\hat l}_n$ and ${\hat s}_n$, Eq.~\ref{eq:3} uses multiplications rather than additions to make the ranges of $\Delta {{\hat l}_n}$ and $\Delta {{\hat s}_n}$ predicted by the CNN less affected by the object size. In addition, the refined OBB predictions are also supervised by the ground truth during the CNN training. Compared to RepPoints \cite{yang2019reppoints}, Oriented RepPoints \cite{li2022oriented}, etc., which first obtain the set of dynamic sampling points and then generate bounding boxes from the point set, the sampling positions of LS-Conv are obtained from OBBs directly. Thus, the receptive field of LS-Conv is always adapted to the object size. It makes the supervision of localization feature sampling in CNN more comprehensive to the final task objective.

\subsection{Convolution for Sampling Classification Features}
Unlike localization feature-sensitive regions directly associated with the OBB representation and spatial coordinates, sensitive regions of classification features are more variable for objects with different categories, shapes, and orientations\cite{9669124,9733170}. Therefore, the sampling positions of the classification convolution are only constrained within the OBB for more flexible adjustment according to each object, as shown in Fig.~\ref{fig:4} (b). Since the OBB is an arbitrary convex quadrilateral, the constraint range is approximated as the minimum external rectangle (MERect)\footnote[2]{OpenCV and many computer vision libraries have implemented this function, so the details are not repeated here.} of the OBB in this case in order to facilitate parallel computation in CNN. Define the length of MERect’s long side as ${S_1}$, the length of the other side as ${S_2}$, the center point of MERect as $\left( {{x_{c}},{y_{c}}} \right)$, and the angle between the long side and the positive direction of $x$-axis as $\alpha$, $\alpha \in \left[ 0, \pi \right)$ in this case. Define the sampling positions of the CS-Conv as ${p_i},i = 0,1, \cdots ,8,$ and the CNN-learnable variables $\omega _i^{\left( x \right)},\omega _i^{\left( y \right)} \in \left( {0,1} \right),{\rm{ }}i = 0,1, \cdots ,8,$ used to adjust the sampling positions. According to the designed constrain, the coordinates of the sampling positions are represented as
\begin{equation}
	\left[ \setlength{\arraycolsep}{0.5pt}{{\begin{array}{*{5}{c}}
			{{x_{{p_i}}}}\\
			{{y_{{p_i}}}}
	\end{array}}} \right] \!= \! \left[ \setlength{\arraycolsep}{3pt}{{\begin{array}{*{5}{c}}
	{\cos \alpha }&{ -\sin \alpha }\\
	{\sin \alpha }&{\cos \alpha }
\end{array}}} \right] \! \times \! \left[\setlength{\arraycolsep}{2pt}{ {\begin{array}{*{5}{c}}
			{{x_c} - 0.5{S_1} + \varpi _{{p_i}}^{\left( x \right)}{S_1}}\\
			{{y_c} - 0.5{S_2} + \varpi _{{p_i}}^{\left( y \right)}{S_2}}
	\end{array}}} \right],
\label{eq:4}
\end{equation}
where $\left( {{x_{{p_i}}},{y_{{p_i}}}} \right)$ denote the coordinates of the sampling position ${p_i},i = 0,1, \cdots ,8$.
\begin{figure}[tbp]
	\centering
	\epsfig{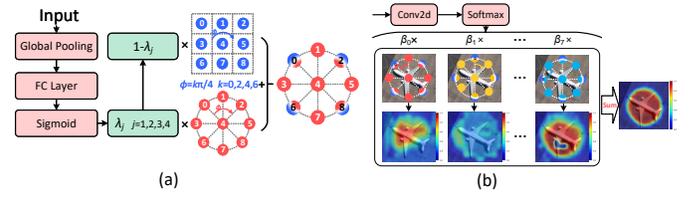}
	\caption{The designed dynamic circluar kernel (DCK). (a) Adaptive fusion of circular and square convolutional kernels. (b) Adaptive fusion of eight-rotation features.}
	\label{fig:5}
\end{figure} 
Although the above design allows the sampling positions to translate to the sensitive regions within the OBB adaptively, the sensitive regions are variable due to the arbitrary orientations of objects in AOOD. It brings more significant uncertainty in the sampling position offsets, and CS-Conv lacks absolute-coordinate information like SCE in LS-Conv. Compared to rotating the sampling positions with more complex constraints, rotating the convolution kernel to the appropriate direction is a more straightforward idea. Thus, a dynamic circular kernel (DCK) is designed. 

First, as shown in Fig.~\ref{fig:5} (a), a circular kernel ${\boldsymbol{\dot K}^{cls}} \in \mathbb{R}^{3\times3}$,
is generated from the square kernel ${\boldsymbol{K}^{cls}} \in \mathbb{R}^{3\times3}$ by bilinear interpolation to accommodate rotations in different directions. The coefficients of ${\boldsymbol{\dot K}^{cls}}$ are defined as ${\dot K}_j^{cls},{\rm{ }}j = 0,1, \cdots ,8$, and the coefficients of ${\boldsymbol{K}^{cls}}$ are defined as ${K}_j^{cls},{\rm{ }}j = 0,1, \cdots ,8$. When $j=1,3,4,5,7$, $\dot K_j^{cls} = K_j^{cls}$. Second, the circular convolution kernel ${\boldsymbol{\dot K}^{cls}}$ is rotated by the angles $\varphi  = \frac{{k\pi }}{4},k = 0,1,2, \cdots ,7,$ in the clockwise direction to obtain eight circular kernels, denoted as ${\boldsymbol{\dot K}_{{{{k\pi } \over 4}}}^{cls}}$. Similarly, the square kernel ${\boldsymbol{K}^{cls}}$ is rotated clockwise by the angles $\varphi  = \frac{{k\pi }}{4},k = 0,2,4,6,$ to get ${\boldsymbol{K}_{{{{k\pi } \over 4}}}^{cls}}$. When $\varphi  = \frac{{k\pi }}{4},k = 0,2,4,6,$ the square kernels and circular kernels are fused to enhance the model fitting ability of CNNs, using the idea of CondConv \cite{yang2019condconv}. The fused kernels are represented as
\begin{equation}
	\boldsymbol{\tilde K}_{{{{k\pi } \over 4}}}^{cls} = \left\{ \setlength{\arraycolsep}{1pt}{\begin{array}{l}
			{\lambda _j}{\boldsymbol{\dot K}_{{{{k\pi } \over 4}}}^{cls}} + \left({1-{\lambda_j}} \right){\boldsymbol{ K}_{{{{k\pi } \over 4}}}^{cls}}, {\quad \rm{if}}\ k = 0,2,4,6,\\
			{\boldsymbol{\dot K}_{{{{k\pi } \over 4}}}^{cls}}, {\quad \rm{if}}\ k  = 1,3,5,7,
	\end{array}} \right.
	\label{eq:6}
\end{equation}
where ${\lambda _j} \in \left( {0,1} \right),j = 1,2,3,4$, denote the CNN-learnable weights for fusion, whose generation module in CNN is shown in Fig.~\ref{fig:5} (a). Similarly, the coefficients of $\boldsymbol{\tilde K}_{{{{k\pi } \over 4}}}^{cls}$ are defined as ${\tilde K}_{j,{{k\pi } \over 4}}^{cls},\ {\rm{ }}j,k = 0,1, \cdots ,8$.

Similar to the localization, define the nine elements (positions) of the following set 
\begin{equation}
\begin{array}{l}
	\left\{ {\left( { - {\textstyle{{\sqrt 2 } \over 2}}, - {\textstyle{{\sqrt 2 } \over 2}}} \right),\left( { - 1,0} \right)} \right.,\left( { - {\textstyle{{\sqrt 2 } \over 2}},{\textstyle{{\sqrt 2 } \over 2}}} \right),\left( {0, - 1} \right),\\
	\left( {0,0} \right),\left( {0,1} \right),\left. {\left( {{\textstyle{{\sqrt 2 } \over 2}}, - {\textstyle{{\sqrt 2 } \over 2}}} \right),\left( {1,0} \right),\left( {{\textstyle{{\sqrt 2 } \over 2}},{\textstyle{{\sqrt 2 } \over 2}}} \right)} \right\}\\
	\buildrel \Delta \over = \left\{ {\left( {\Delta x_v^{cls},\Delta y_v^{cls}} \right),v = 0,1, \cdots ,8} \right\}.
\end{array}
\label{eq:0_2}
\end{equation}
The indices of $\left( {x,y} \right)$ and its eight circular neighbors are represented as $\left( {x + \Delta x_v^{cls},{\rm{ }}y + \Delta y_v^{cls}} \right)$, $v = 0,1, \cdots ,8$. Define the feature map input to the classification branch as ${\boldsymbol{I}^{cls}} \in {\mathbb{R}^{W \times H \times F}}$. The elements of ${\boldsymbol{I}^{cls}}$ at $\left( {x + \Delta {x_v^{cls}},{\rm{ }}y + \Delta {y_v^{cls}}} \right)$, $v = 0,1, \cdots ,8$, are defined as $\boldsymbol{I}_{x + \Delta {x_v^{cls}},y + \Delta {y_v^{cls}}}^{cls}$, which are $F$-dimensional feature vectors. When position $\left( {x,y} \right)$ is a positive position, as shown in Fig.~\ref{fig:4} (d), the nine sampling points $\left( {x + \Delta {x_v^{cls}},{\rm{ }}y + \Delta {y_v^{cls}}} \right),{\rm{ }}v = 0,1, \cdots ,8,$ are correspondingly moved to the points $\left( {{x_{{{p}_i}}},{y_{{{p}_i}}}} \right),\ {\rm{  }}i = 0,1, \cdots ,8,$ by DCN \cite{dai2017deformable}. The sampled feature vectors are denoted as $\boldsymbol{I}_{{x_{{p_i}}},{y_{{p_i}}}}^{cls},{\rm{ }}i = 0,1, \cdots ,8$. When the position $\left( {x,y} \right)$ is not a positive position, the sampling points $\left( {x + \Delta {x_v^{cls}},{\rm{ }}y + \Delta {y_v^{cls}}} \right),\ {\rm{ }}v = 0,1, \cdots ,8,$ do not move. Thus, the output of CS-Conv at $\left( {x,y} \right)$ is represented as
\begin{equation}
	\! \boldsymbol{O}_{x,y}^{cls} \! = \! \left\{ \begin{array}{l}
		\! \sum\limits_{k = 0}^{M = 7} {\sum\limits_{\! i, j = 0}^{\! N = 8} { {\boldsymbol{I}_{{x_{{p_i}}},{y_{{p_i}}}}^{cls}\! {\tilde K}_{j,{{k\pi} \over 4}}^{cls} } } {\beta _k} m_j^{cls}},{\enspace \rm{if}}\ (x,y){\ \rm{ is \ positive}},\\
	\! \sum\limits_{k = 0}^{M = 7} {\! \sum\limits_{v, j = 0}^{N = 8} {\boldsymbol{I}_{x + \Delta {x_v^{cls}},y + \Delta {y_v^{cls}}}^{cls}\! {\tilde K}_{j,{{k\pi} \over 4}}^{cls} {\beta _k} m_j^{cls}}} ,{\enspace \rm{otherwise,}}
	\end{array} \right.
	\label{eq:7}
\end{equation}
where $\boldsymbol{O}_{x,y}^{cls} \in \mathbb{R}^{1\times1\times F}$, and CNN-learnable scalars ${\beta _k},k = 0,1, \cdots ,7,{\rm{ }}\sum\nolimits_{k = 0}^{M = 7} {{\beta _k} = 1} ,$ are used to re-weight the features extracted by ${\boldsymbol{\tilde K}_{{{{k\pi } \over 4}}}^{cls}}$ with different orientations, as shown in Fig.~\ref{fig:5} (b). CNN-learnable scalars $m_j^{cls} \in \left( {0,1} \right)$, $j = 0,1, \cdots ,8,$ are utilized to adjust the contributions of different sampling positions. In CNN, CS-Conv with multiple orientations is implemented by group convolution to save computational costs.

\subsection{Dynamic Task-consistent-aware Label Assignment}
\begin{figure}[tb]
	\centering
	\epsfig{width=0.48\textwidth,file=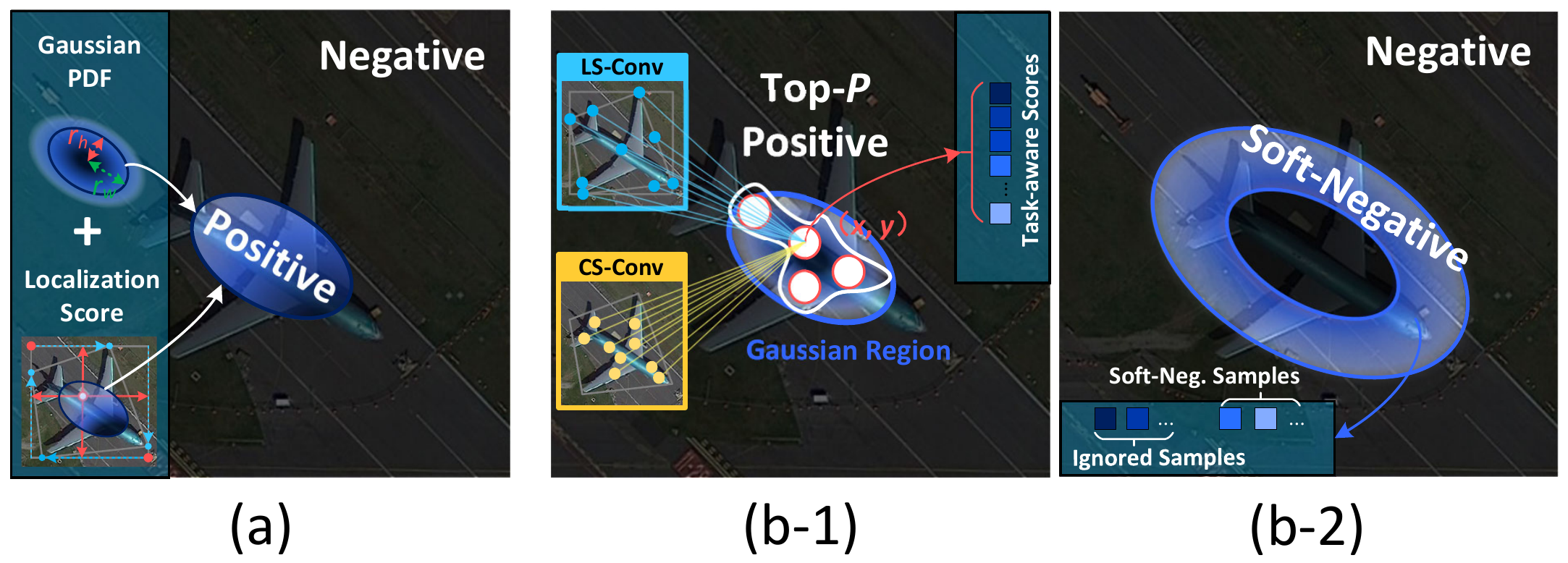}
	\caption{The label assignment strategy for AOOD: (a) GGHL \cite{huang2022general} and (b) DTLA. The designed DTLA consists of (b-1) the dynamic positive candidate position assignment based on task-consistent-aware scores and (b-2) the soft-weighted negative candidate position assignment.}
	\label{fig:6}
\end{figure}

For the CNN-based object detection, an object may correspond to more than one candidate detection position in the feature maps. It is significant to select the appropriate positive and negative positions from them to assign labels for CNN training. The label assignment strategy has been studied in many works such as ATSS \cite{zhang2020bridging}, Autoassign \cite{zhu2020autoassign}, etc. In the field of arbitrary-oriented object detection, GGHL \cite{huang2022general} selected candidate positions based on Gaussian heatmaps and OBB prediction scores, as shown in Fig.~\ref{fig:6} (a). However, on one hand, it uses a hard-thresholding selection strategy, i.e., the candidate positions with the value of Gaussian heatmap score $F_{x,y}$ higher than the threshold $T$ are positive, and the other positions are negative. See Appendix B for the derivation of $F_{x,y}$. The fixed candidate regions may lead to misassignment due to the irregularity and variety of object shapes. On the other hand, a location closer to the center of the Gaussian heatmap may not be a better positive position. Only using localization score to adjust the weights of positive positions in GGHL \cite{huang2022general} still faces the IFS problem that these positive positions may not be optimal for both localization and classification tasks. 
\begin{algorithm}[!tbp]
	{\label{alg:1}
	\scriptsize
	\caption{DTLA}
	\LinesNumbered 
	\KwIn{The encoded ground truth $\boldsymbol{l}_{x,y}$, $\boldsymbol{s}_{x,y}$, ${ar}_{x,y}$, $C^{(h)}_{x,y}$, ${{\widehat {obj}}_{x,y}}$. The CNN predictions $\tilde{\boldsymbol{l}}_{x,y}$, $\tilde{\boldsymbol{s}}_{x,y}$, $\tilde{{ar}}_{x,y}$, $\hat{C}^{(h)}_{x,y}$, ${{\widehat {obj}}_{x,y}}$. $F_{x,y} \in R_{gh}$ obtained by static label assignment of GGHL. The prior threshold $T$. The number of object $N_{obj}$}
	\KwOut{The objectness loss $Loss^{obj}$}
	\eIf{${F_{x,y}} > 0$}{
		Calculate the localization score $L_{x,y}$ by Eqs. (\ref{eq:9}-\ref{eq:10}) \;
		Calculate the combined score ${D_{x,y}}$ by Eq. (\ref{eq:8}) \;
		\For{$ i \in N_{obj}$}{	
			\For{$(x,y) \in R^{i}_{gh}$}{
				\eIf{${F_{x,y}} > T$}{
					$P = \left\lceil {\sum\limits_{\left( {x,y} \right) \in {R_{gh}}} {{L_{x,y}}} } \right\rceil $\;
					Select the Top-$P$ ${{D}_{x,y}}$\;
					\eIf{${{D}_{x,y}}$ in Top-$P$ ${{D}_{x,y}}$}{$(x,y) \in R_{pos}$;}
					{$(x,y) \in R_{ig}$;}
				}{
					\eIf{${{D}_{x,y}} < T$}{$(x,y) \in R_{sneg}$ \;
						${w_{sneg}} = 1 - {D_{x,y}}$ \;	
					}{$(x,y) \in R_{ig}$}
				}
			}
		}
	}{$(x,y) \in R_{neg}$}
	Calculate the objectness loss $Loss^{obj}$ by Eq. (\ref{eq:12}).}
\end{algorithm}
In response, DTLA strategy is designed based on TS-Conv and GGHL \cite{huang2022general}. DTLA divides the regions of candidate positions into positive, negative, soft-negative, and ignored regions denoted as the position sets $R_{pos}$, $R_{neg}$, $R_{sneg}$, and $R_{ig}$. These regions are dynamically adjusted according to the localization and classification prediction costs to obtain the optimal position combinations for CNN training. The detailed implementation of DTLA is given in Algorithm~\ref{alg:1}.

\textbf{\textit{1) Positive positions.}} Different from GGHL \cite{huang2022general}, which treats all the positions in a fixed range of a Gaussian ellipse region as positive positions, the proposed DTLA ranks the candidate positions of each object and takes the Top-$P$ positions as candidate positions. Since the performance of AOOD is usually evaluated by the product of localization and classification scores, the straightforward idea is to rank the positive candidate positions according to the combined score ${D_{x,y}}$ of localization and classification.

First, the static arbitrary-oriented label assignment strategy GGHL \cite{huang2022general} is utlized as the initial assignment of DTLA. Based on the initial assignment, the designed dynamic label assignment strategy is carried out. If a position ${D_{x,y}}$ lies in the Gaussian region, $F_{x,y} > T$, and the ranking of ${D_{x,y}}$ is within the Top-$P$, then this position is positive for assigning the label to predict the object, $\left( {x,y} \right) \in {R_{pos}}$. In this case, we set $T=0.3$. Benefiting from the designed TS-Conv, localization and classification features are extracted from their sensitive regions respectively and mapped to the same position, as shown in Fig.~\ref{fig:6} (b). The optimal task-wise scores are spatially aligned for determining positive candidate positions. Thus, the localization and classification combined score at $\left( {x,y} \right)$ is defined as
\begin{equation}
	\! {D}_{x,y} \! = \left\{ \setlength{\arraycolsep}{0.5pt}{\begin{array}{l}
			\tilde \vartheta F_{x,y} \! + \! (1 \! - \! \tilde \vartheta){\! \sqrt {\! {L_{x,y}}{\hat C}_{x,y}^{\left( h \right)}} }{\quad \rm{if}} (x,y) \! \in \! {R_{gh}}\\
			{0} {\quad \rm{otherwise}}
	\end{array}} \right.,
	\label{eq:8}
\end{equation}
where ${\hat C}_{x,y}^{\left( h \right)} \in \left( {0,1} \right)$ indicates the predicted classification score that the object belongs to the ground truth category $h$ and will be specified later. The localization score is represented as
\begin{equation}
{L_{x,y}} = {e^{ - Loss_{x,y}^{loc}}},
	\label{eq:9}
\end{equation}
\begin{equation}
\begin{array}{l}
	Loss_{x,y}^{loc} = 1 - GIoU\left( {{\boldsymbol{l}_{x,y}},{{\tilde{\boldsymbol{l}}}_{x,y}}} \right)\\
	+ MSE\left( {{\boldsymbol{s}_{x,y}},{{\tilde{\boldsymbol{s}}}_{x,y}}} \right) + {\left( {{a_{x,y}} - {{\tilde a}_{x,y}}} \right)^2}
\end{array},
	\label{eq:10}
\end{equation}
where $Loss_{x,y}^{loc}$ denotes the OBB localization loss at $\left( {x,y} \right)$. $L_{x,y} \in (0,1)$ is monotonically decreasing with $Loss_{x,y}^{loc}$. ${\tilde{\boldsymbol{l}}_{x,y}} = \left[ {{{\tilde l}_1},{{\tilde l}_2},{{\tilde l}_3},{{\tilde l}_4}} \right]$ and ${\tilde{\boldsymbol{s}}_{x,y}} = \left[ {{{\tilde s}_1},{{\tilde s}_2},{{\tilde s}_3},{{\tilde s}_4}} \right]$ represent the predictions of an OBB based on the output localization features $\boldsymbol{O}_{x,y}^{loc}$ of TS-Conv. ${\boldsymbol{l}_{x,y}} = \left[ {{l_1},{l_2},{l_3},{l_4}} \right]$ and ${\boldsymbol{s}_{x,y}} = \left[ {{s_1},{s_2},{s_3},{s_4}} \right]$ represent the ground truth of this OBB. ${a_{x,y}}$ and ${\tilde a_{x,y}}$ denote the ground truth and predictions of the area ratio of the OBB and HBB, respectively. $GIoU\left( . \right)$ is the function to measure the HBB localization accuracy by Generalized Intersection over Union (GIoU) \cite{rezatofighiGeneralizedIntersectionUnion2019}. $MSE\left(  \cdot  \right)$ represents the mean square error (MSE) function. See GGHL \cite{huang2022general} for details of $Los{s^{loc}}\left( {x,y} \right)$. The hyperparameter $\tilde \vartheta$ is defined as
\begin{equation}
\tilde \vartheta  = \frac{{ite{r_{\max }} - iter}}{{ite{r_{\max }}}} \times \vartheta,
\label{eq:11}
\end{equation}
where $iter$ denotes the number of current iterations and $iter_{max}$ denotes the number of maximum iterations during CNN training. Here, $\vartheta  = 0.3$ according to our empirical study. In the initial stage of CNN training, the label assignment mainly relies on the prior Gaussian heatmaps generated by GGHL \cite{huang2022general} due to inaccurate localization and classification. As the CNN training converges, the label assignment becomes more dependent on the increasing score. The variable $P = \left\lceil {\sum\limits_{\left( {x,y} \right) \in {R_{gh}}} {{L_{x,y}}} } \right\rceil $, where $\left\lceil . \right\rceil $ denotes the upward rounding operation, is dynamically adjusted for different objects. The positive positions are no longer fixed but dynamically selected according to ${D_{x,y}}$. 

The above strategy takes full advantage of the proposed task-wise sampling convolutions that can extract and map the most sensitive features of different tasks to the same spatial location. On one hand, it helps to select the optimal candidate positions suitable for both localization and classification dynamicly according to different objects in different scenes. On the other hand, the low-quality candidate positions due to MERect approximation can be filtered according to DTLA. Thus, it is unnecessary to generate the Gaussian probability density function (PDF) of any convex quadrilateral, but only needs to replace it with the Gaussian PDF based on MERect, which makes the algorithm more efficient and concise.

\textbf{\textit{2) Negative positions.}} The designed DTLA uses a soft thresholding strategy instead of treating all positions with ${F_{x,y}} < T$ as negative, as shown in Fig.~\ref{fig:6} (b). If $\left( {x,y} \right)$ does not lie in the Gaussian region, this position is negative, and the background label is assigned, $\left( {x,y} \right) \in {R_{neg}}$. If $\left( {x,y} \right)$ lies in the Gaussian region $R_{gh}$ but ${F_{x,y}} < T,{\rm{ }} \ {D_{x,y}} < T$, this position is considered as soft negative position, $\left( {x,y} \right) \in {R_{sneg}}$. The background prediction loss of these positions is multiplied by the weight ${w_{sneg}} = 1 - {D_{x,y}},{\rm{ }}{w_{sneg}} \in \left( {0,1} \right)$. The smaller the ${D_{x,y}}$, the larger the weight, indicating a higher negative attribute for this position.

\textbf{\textit{3) Ignored positions.}} If $\left( {x,y} \right)$ lies in the Gaussian region $R_{gh}$, ${F_{x,y}} > T$, but ${D_{x,y}}$ is not within the Top-$P$, this position may not satisfy both localization and classification tasks, although it have a high priori score. If $\left( {x,y} \right)$ lies in the Gaussian region $R_{gh}$, ${F_{x,y}} < T$, but  ${D_{x,y}} > T$, this position is too close to the junction region between the object and the background for a low a priori score, although it obtains a high ${D_{x,y}}$. In the above two cases, the priori score ${F_{x,y}}$ and the dynamic score ${D_{x,y}}$ contradict each other. It is not appropriate to treat $\left( {x,y} \right)$ as either positive or negative position, so it is ignored and not used for CNN training, i.e., $\left( {x,y} \right) \in {R_{ig}}$.

\textbf{\textit{4) Loss functions.}} If a position $\left( {x,y} \right)$ is positive, its assigned ground truth of objectness $ob{j_{x,y}} = L_{x,y}$; if $\left( {x,y} \right)$ is negaitive, $ob{j_{x,y}} = 0$. Note that, the CNN gradient of ${L_{x,y}}$ is not backpropagated. Define the CNN predicted objectness score at $\left( {x,y} \right)$ as ${\widehat {obj}_{x,y}} \in \left( {0,1} \right)$. According to the assignments of different candidate positions, the binary loss function for objectness prediction is represented as
\begin{equation}
	\setlength{\arraycolsep}{0.3pt}{\begin{array}{l}
		Los{s^{obj}} \! =  \! -\frac{1}{{{M_{pos}}}} \! \sum\limits_{\tiny{(x,y) \in {R_{pos}}}}\! {{{\left| {L_{x,y} - {{\widehat {obj}}_{x,y}}} \right|}^\gamma }\log  {{{\widehat {obj}}_{x,y}}} } \\
		- \frac{1}{{{M_{neg}}}} \sum\limits_{(x,y) \in {R_{neg}}} {{{{{\widehat {obj}}_{x,y}}}^\gamma }\log \left( {{{ {1 - \widehat{obj}}}_{x,y}}} \right)} \\
		- \frac{1}{{{M_{sneg}}}} \sum\limits_{(x,y) \in {R_{sneg}}} {{w_{sneg}}{{{{\widehat {obj}}_{x,y}}}^\gamma }\log \left( {{{ {1 - \widehat{obj}}}_{x,y}}} \right)} 
	\end{array}},
	\label{eq:12}
\end{equation}
where ${M_{pos}}$, ${M_{neg}}$, and ${M_{sneg}}$ mean the numbers of positive, negative, and soft-negative positions for an input image, respectively. The hyperparameter of Focal Loss \cite{linFocalLossDense2017} $\gamma = 2$, which is the same as GGHL \cite{huang2022general}. The OBBs and category labels of the objects are assigned to positive positions filtered by TOP-$P$ strategy for supervised CNN predictions for object localization and classification. According to Eq.(\ref{eq:10}), the localization loss is calculated as
\begin{equation}
	Los{s^{loc}} = \frac{1}{{{M_{pos}}}} \times \sum\limits_{\left( {x,y} \right) \in {R_{pos}}} {\left( {Loss_{x,y}^{init} + Loss_{x,y}^{loc}} \right)},
	\label{eq:13}
\end{equation}
which represents the OBB localization loss of the initial stage, which is used to suprivise the sampling points of TS-Conv.

Define the assigned ground truth of the $h$th category at $\left( {x,y} \right)$ as $C_{x,y}^{\left( h \right)}$. If the object at $\left( {x,y} \right)$ belongs to the $h$th category, $C_{x,y}^{\left( h \right)} = 1$; otherwise, $C_{x,y}^{\left( h \right)} = 0$. Define the CNN prediction of $C_{x,y}^{\left( h \right)}$ as ${\hat C}_{x,y}^{\left( h \right)} \in \left( {0,1} \right)$. The classification loss is calculated as
\begin{equation}
\setlength{\arraycolsep}{0.3pt}{\begin{array}{l}
	Los{s^{cls}} = \frac{1}{{{M_{pos}}}}  \! \times \! \sum\limits_{\left( {x,y} \right) \in {R_{pos}}}  \! {\sum\limits_{h = 1}^{{M_C}} {\left( {C_{x,y}^{\left( h \right)}\log \left( {\hat C_{x,y}^{\left( h \right)}} \right)} \right.} } \\
	\left. { + \left( {1 - C_{x,y}^{\left( h \right)}} \right)\log \left( {1 - \hat C_{x,y}^{\left( h \right)}} \right)} \right)
\end{array}},
	\label{eq:15}
\end{equation}
where  ${M_C}$ represents the total number of categories of objects. The total loss is stated as, 
\begin{equation}
Loss = Los{s^{obj}} + Los{s^{loc}} + Los{s^{cls}},
	\label{eq:16}
\end{equation}
which represents the sum of the objectness loss, localization loss and classification loss.

%% file: Chap4.tex
In this section, extensive experiments are conducted on several AOOD datasets to evaluate the performance of TS-Conv comprehensively. First, the experimental conditions, including datasets, evaluation metrics, implementation details, etc., are introduced. Then, ablations experiments are designed to validate the effectiveness of each component of the TS-Conv model and to evaluate their performance quantitatively. The task-wise samplings, rotation-invariant feature extraction, dynamic label assignment, and other previously claimed issues in AOOD are discussed in detail. Forthermore, the scalability of TS-Conv on lightweight models and for multimodal data are evaluated. After that, comparative experiments with existing AOOD methods on the datasets covering different scenarios are given and analyzed. Besides, the extension of the proposed TS-Conv to improve the performance of lightweight AOOD models on embedded devices is further explored.

\subsection{Experimental Conditions}
\textbf{1) \textit{Experimental platforms.}} The experiments are performed on a server with an AMD 3950WX CPU, 128 GB memory, and four NVIDIA GeForce RTX 3090 GPU (24GB). In addition, the performance of the TS-Conv improved lightweight models are evaluated on embedded edge devices NVIDIA Jetson AGX Xavier and Jetson TX2.

\textbf{2) \textit{Datasets.}} To evaluate the performance of TS-Conv model more comprehensively, several public AOOD datasets are used, which cover different scenes, different shapes and categories of objects, and different data sources.

a) DOTAv1.0 \cite{xiaDOTALargeScaleDataset2018} dataset provides a widely used benchmark to evaluate the performance of AOOD methods in remote sensing scenes. It has more than 188,000 objects covering 15 categories in 2,806 images from $800\times800$ pixels to $4,000\times4,000$ pixels. Due to the huge size of remote sensing images, they are usually cropped into sub-images of $800\times800$ pixels with an overlap of 200 pixels on each dimension after being scaled to different sizes with the ratios of 0.5, 1.0, and 1.5 \cite{huang2021lo}. DOTAv2.0 \cite{li2020object} further expands the number of objects to 1,793,658 objects covering 18 categories. 

b) HRSC2016 \cite{liu2017high} is a ship detection dataset consisting of 436 training images, 181 validation images, and 444 testing images from $300\times300$ pixels to $1500\times900$ pixels.

c) DIOR-R \cite{ding2021object,cheng2022anchor} dataset is an aerial AOOD datasets contains 190,288 objects covering 20 categories in 23,463 images with the size of $800\times800$ pixels. In the DIOR-R dataset, 5,862 images are used for training, 5,863 images are used for validation, and 11,738 images are used for testing. 

d) DroneVehicle \cite{sun2022drone} is an infrared-RGB vehicle detection datasets. After screening and pre-processing, 17,900 Infrared-RGB image pairs are used for training, 1,469 Infrared-RGB image pairs are used for validation, and 8980 Infrared-RGB image pairs are used for testing.

e) SSDD+ \cite{li2017ship} is a synthetic aperture radar (SAR) dataset for ship detection. It has 1,160 ship images including 2,456 instances collected from different sea conditions. The ratio of training, validation and testing images is 7:1:2.

\textbf{3) \textit{Evaluation metrics.}} The mean Average Precision (mAP) is adopted for evaluating the detection accuracy. The mAP with an IoU threshold of 50\% is represented as mAP$_{50}$. mAP$_{50:95}$ means calculating the mAP when the IoU threshold is 50\% to 95\% at every 5\% interval and then taking their mean as mAP$_{50:95}$. The speed is evaluated by frames per second (fps) and the computational complexity is evaluated by the floating point operations (FLOPs) for lightweight models.

\textbf{4) \textit{Implementation details.}}  The initial learning rate for training DOTA, DIOR-R, HRSC2016 and SSDD+ datasets is $5\times10^{-4}$, and the final learning rate is $1\times10^{-6}$. The optimizer for training the CNN model is stochastic gradient descent (SGD), and the learning rate scheduler is cosine decay. The weight decay and momentum are set as $5\times10^{-4}$ and 0.9, respectively. The batch size is 32 (8 images per GPU). The maximum training epoch for DOTA, DIOR-R, HRSC2016, DroneVehicle and SSDD+ datasets are 36, 150, 150, 50 and 150, respectively. The non-maximum suppression (NMS) threshold is 0.4. Random cropping, random flipping, random rotation and mixup strategies are employed for data augmentation. The experiments of the proposed TS-Conv on the DOTA datasets use multi-scale cropped images as the training set but use single-scale cropped images as the testing set. The multi-scaling data augmentation is not used during model training and testing. The experiments of the proposed TS-Conv on other datasets all adapt single-scale training and single-scale testing strategies.

\subsection{Ablation Experiments}
The results of the ablation experiments for different components of TS-Conv, including LS-Conv, CS-Conv with DCK, and DTLA, are presented in Table~\ref{table:2}. The anchor-free AOOD method GGHL \cite{huang2022general} is chosen as the baseline, and experiments are conducted on the most widely used AOOD dataset DOTA v1.0 \cite{xiaDOTALargeScaleDataset2018}. 

\textbf{1) Ablation experiments of each component.} First, as a control group, the shared-offset DCNs employed by RepPoints \cite{yang2019reppoints}, Oriented RepPoints \cite{li2022oriented}, etc., are introduced based on GGHL. This design has the same sampling positions for localization and classification features and maps these features to aligned locations. Although the above scheme somewhat enables more objects to be detected and improves the mAP$_{50}$ by 0.82, the improvement of the stricter metrics, i.e., mAP$_{75}$ and mAP$_{50:95}$, is slight. It implies that the model may relax its requirements for detection quality to accommodate two different tasks.
\begin{table*}[tp]
	\centering
	\renewcommand\arraystretch{1.2}
	\setlength{\tabcolsep}{2.5mm}{
		\caption{\label{table:2}
			{Ablation experiments of the proposed TS-Conv on the DOTAv1.0 dataset}}
		\resizebox{\textwidth}{!}{\setlength{\tabcolsep}{1.5mm}{
				\begin{tabular}{c|c|c|c|c|c|c|c|cccc}
					\hline\hline
					{\multirow{2}{*}{Methods}}     & \multirow{2}{*}{\begin{tabular}[c]{@{}c@{}}Decoupled \\ Head\end{tabular}} &\multirow{2}{*}{\begin{tabular}[c]{@{}c@{}} DCNs  \end{tabular}} &
					\multicolumn{3}{c|}{\begin{tabular}[c]{@{}c@{}}Task-wise DCNs \\ (TS-DCN) \end{tabular}}& \multicolumn{2}{c|}{\begin{tabular}[c]{@{}c@{}}Label \\ Assignment \end{tabular}} &  \multicolumn{3}{c}{mAPs on the DOTAv1.0 Dataset} & \multirow{2}{*}{\begin{tabular}[c]{@{}c@{}}Speed \\ (fps)\end{tabular}}\\ \cline{4-11}
					& & & LS-Conv & CS-Conv & DCK & GGHL & DTLA & mAP$_{50}$ & mAP$_{75}$ & mAP$_{50:95}$ \\ \hline
					
					Baseline (GGHL \cite{huang2022general}) & $\checkmark$ & & & & & $\checkmark$ &  & 76.95 & 44.19 & 44.29 & 42.30 \\ \hline

					\begin{tabular}[c]{@{}c@{}}TOOD + DCNs \cite{9710724}\end{tabular} & T-head & & & & & TAL+TAP improved &  & 77.59 \tiny{(+0.64)} & 44.68 \tiny{(+0.49)} & 44.64 \tiny{(+0.35)} & 23.23 \\ \hline
					
					\begin{tabular}[c]{@{}c@{}}Shared-offset DCNs\end{tabular} & $\checkmark$ &  shared-offsets & & & & $\checkmark$ &  & 77.77 \tiny{(+0.82)} & 44.76 \tiny{(+0.57)} & 44.99 \tiny{(+0.70)} & 23.23 \\ \hline
					
					\begin{tabular}[c]{@{}c@{}}DCNs without shared-offsets \end{tabular} & $\checkmark$ &   w/o shared-offsets & & & & $\checkmark$ &  & 77.39 \tiny{(+0.44)} & 44.52 \tiny{(+0.33)} & 44.60 \tiny{(+0.31)} & 23.23 \\ \hline
					
					LS-Conv& $\checkmark$ & & $\checkmark$ & & & $\checkmark$ &  & 78.08 \tiny{(+1.13)} & 46.38 \tiny{(+2.19)} & 45.87 \tiny{(+1.27)} & 31.08\\ \hline
					
					\begin{tabular}[c]{@{}c@{}} CS-Conv \end{tabular} & $\checkmark$ & & & $\checkmark$ &  & $\checkmark$ &  & 77.84 \tiny{(+0.89)} & 44.23 \tiny{(+0.04)} & 45.37 \tiny{(+1.08)} & 31.08\\ \hline 
					
					\begin{tabular}[c]{@{}c@{}}CS-Conv with DCK\end{tabular}& $\checkmark$ & & &  $\checkmark$ & $\checkmark$ & $\checkmark$ &  & 78.06 \tiny{(+1.11)} & 45.46 \tiny{(+1.27)} & 45.36 \tiny{(+1.07)} & 28.96 \\ \hline
					
					DTLA & $\checkmark$ & & & & & & $\checkmark$ & 77.95 \tiny{(+1.00)} & 46.10 \tiny{(+1.91)} & 45.38 \tiny{(+0.70)} & 42.30\\ \hline
					
					TS-DCN & $\checkmark$ & & $\checkmark$ & $\checkmark$ & $\checkmark$ & $\checkmark$ &  & 78.13 \tiny{(+1.18)} & 46.59 \tiny{(+2.40)} & 46.09 \tiny{(+1.80)} & 23.23\\ \hline
					
					\textbf{TS-Conv} & $\checkmark$ & & $\checkmark$ & $\checkmark$ & $\checkmark$ & & $\checkmark$ & \textbf{78.75 \tiny{(+1.80)}} & \textbf{46.60 \tiny{(+2.41)}} & \textbf{46.27 \tiny{(+1.98)}} & 23.23\\ \hline
					\hline\hline
	\end{tabular}}}}\vspace{0.5em}
	\justifying{Note: Bold indicates the best result. 'Shared-offset DCN' represents that both branches of the decoupled head use DCN \cite{dai2017deformable} and their sampling offsets are shared as in the case of RepPoints \cite{yang2019reppoints} and Oriented RepPoints \cite{li2022oriented}. The convolutions of TOOD's detection head also use DCNs for fair.}
\end{table*}
\begin{figure}[!t]
	\centering
	\epsfig{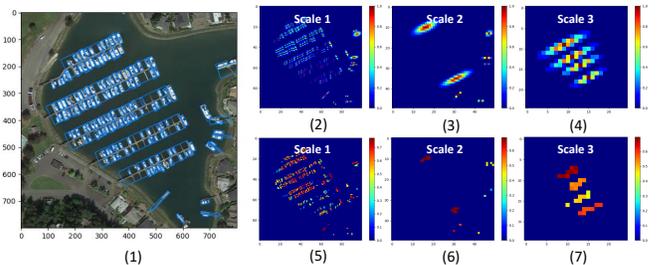}
	\caption{Comparison of DTLA and GGHL label assignment strategies. Figs. (2)-(4) represent the positive candidate positions and their scores statically assigned by the GGHL strategy at three different scales, respectively. Figs. (5)-(7) represent the positive candidate positions and their scores dynamically assigned by the proposed DTLA strategy at three different scales, respectively. The closer the color is to red, the higher the score.}\label{fig:7}
\end{figure}
\begin{figure}[!t]
	\centering
	\epsfig{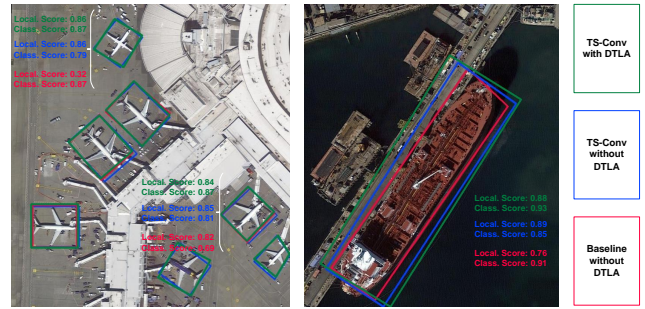}
	\caption{Visualization results of task-wise prediction scores.}\label{fig:AR1}
\end{figure}

\begin{table}[!t]
	\centering
	\renewcommand\arraystretch{1}
	\setlength{\tabcolsep}{3mm}{
		\caption{\label{table:3}
			{Evaluations of detection performance for different values of hyperparameters $T$ and $\vartheta$, and experiments of using DCK and random rotation data augmentation on the DOTAv1.0 dataset}}}
	\resizebox{0.48\textwidth}{!}{\setlength{\tabcolsep}{3mm}{
			\begin{tabular}{cc||cc||ccc}
				\hline\hline
				$T$ & mAP$_{50}$ & $\vartheta$ & mAP$_{50}$ & RandomRota & DCK & mAP$_{50}$ \\	
				\hline
				0.2 & 78.45 & 0.3 & \textbf{78.75} & $\checkmark$ & & 77.92 \\
				0.3 & \textbf{78.75} & 0.4 & 78.52 & & $\checkmark$ & 78.47\\
				0.4 & 78.60 & 0.5 & 78.50 & $\checkmark$ & $\checkmark$ & \textbf{78.75}\\
				\hline\hline
	\end{tabular}}}\vspace{0.5em}
	\justifying{Note: Bold indicates the best result. RadomRota represents using the random rotation data agumentation operation during the CNN training. When evaluating one variable, the other variables are fixed.}
\end{table}

Correspondingly, the proposed task-wise DCNs consisting of LS-Conv and CS-Conv are evaluated. With the introduction of LS-Conv, the mAP$_{75}$ is improved by 4.96\% (+2.19) compared to the baseline and 3.63\% (+1.62) compared to the shared-offset DCNs. TOOD \cite{9710724}, which also adopts task-specific feature learning ideas, is also compared. The results show that TS-Conv, which uses more explicit constraints and considers directional robustness, performs better than TOOD, which uses reweighted implicit feature extraction constraints. The designed LS-Conv further improves the localization accuracy compared to the shared-offset DCNs by directly associating the convolutional sampling points with the OBB representation and embedding the spatial coordinates into the features. When only CS-Conv is used without LS-Conv, i.e., the sampling points of convolutions are just constrained in OBB, the mAP$_{50}$, mAP$_{75}$ and mAP$_{50:95}$ are also increased, but the improvement, especially for mAP$_{75}$, is minor. The bottleneck of this scheme is mainly in inaccurate localization. Because the spatial location coordinates of the OBB are explicit and essential features for the localization task, the convolutional sampling constraint of CS-Conv is vague for localization, although it is suitable for classification. In addition, since the objects in AOOD have richer orientation variations, performance improvements are also observed after replacing the convolutional kernel of CS-Conv with the designed DCK. Furthermore, when LS-Conv and CS-Conv are combined to obtain the TS-DCN scheme, significant performance gains are seen compared to using them alone as listed in Table~\ref{table:2}. The mAP$_{50}$, mAP$_{75}$ and mAP$_{50:95}$ are increased by 1.53\% (+1.18), 5.43\% (+2.40), and 4.06\% (+1.80), respectively. It illustrates that the LS-Conv and CS-Conv play complementary roles for each other. LS-Conv compensates for the lack of localization accuracy using only CS-Conv, while CS-Conv with DCK further improves detection performance.

Then, the performance of the two label assignment strategies, GGHL \cite{huang2022general} and the proposed DTLA, is evaluated. The proposed DTLA increases the mAP$_{50}$, mAP$_{75}$ and mAP$_{50:95}$ without introducing additional CNN structure and adding inference cost. Fig.~\ref{fig:7} shows the positive candidate positions selected by GGHL \cite{huang2022general} and DTLA for predicting objects. As seen in Fig.~\ref{fig:7}, there is a significant difference between the positive candidate positions dynamically selected by DTLA according to ${\tilde D}_{x,y}$ and the positions statically assigned by GGHL \cite{huang2022general} according to the Gaussian prior. The Gaussian peak position is not necessarily the optimal candidate position. It is more reasonable for DTLA to select the positive candidate positions adaptively according to different objects and training stages. Besides, Table~\ref{table:5A} also compares the proposed DTLA with other label assignment strategies used for general object detection. The experimental results show that DTLA brings more benefits compared to other label assignment strategies. Furthermore, Fig.~\ref{fig:AR1} show the visualization results of task-wise prediction scores, from which it can be seen that the proposed TS-Conv with DTLA has better localization and classification prediction consistency.

\begin{table}[tbp]
	\centering
	\renewcommand\arraystretch{1.1}
	\setlength{\tabcolsep}{9mm}{
		\caption{\label{table:5A}
			{Evaluations of different label assignment strategies on the DOTA datasets}}
		\resizebox{0.49\textwidth}{!}{\setlength{\tabcolsep}{4mm}{
				\begin{threeparttable}
					\begin{tabular}{cccc}
						\hline\hline
					
						\multicolumn{1}{c|}{{}}  &  \multicolumn{1}{c|}{{}} & \multicolumn{1}{c|}{{}} & {}  \\
						\multicolumn{1}{c|}{\multirow{-2}{*}{{Models}}}                                            & \multicolumn{1}{c|}{\multirow{-2}{*}{{Label Assign.}}} & \multicolumn{1}{c|}{\multirow{-2}{*}{{mAP$_{50}$}}}                                      & \multirow{-2}{*}{{\begin{tabular}[c]{@{}c@{}}Inference\\ Speed (fps)\end{tabular}}}  \\ \hline
						
						\multicolumn{1}{c|}{{\begin{tabular}[c]{@{}c@{}}GGHL\cite{huang2022general}\end{tabular}}} &  \multicolumn{1}{c|}{{Anchor\cite{redmonYOLOv3IncrementalImprovement2018}}} & \multicolumn{1}{c|}{{ \begin{tabular}[c]{@{}c@{}}74.64 \end{tabular}}} & {\begin{tabular}[c]{@{}c@{}}38.77\\ \end{tabular}}                          \\
						
						\multicolumn{1}{c|}{{\begin{tabular}[c]{@{}c@{}}GGHL\cite{huang2022general}\end{tabular}}} &  \multicolumn{1}{c|}{{Centerness\cite{tian2019fcos}}} & \multicolumn{1}{c|}{{ \begin{tabular}[c]{@{}c@{}}73.48 \end{tabular}}} & {\begin{tabular}[c]{@{}c@{}}42.39\\ \end{tabular}}                          \\
						
						\multicolumn{1}{c|}{{\begin{tabular}[c]{@{}c@{}}GGHL\cite{huang2022general}\end{tabular}}} &  \multicolumn{1}{c|}{{ATSS\cite{zhang2020bridging}}} & \multicolumn{1}{c|}{{ \begin{tabular}[c]{@{}c@{}}75.15 \end{tabular}}} & {\begin{tabular}[c]{@{}c@{}}39.06\\ \end{tabular}}                          \\
						
						\multicolumn{1}{c|}{{\begin{tabular}[c]{@{}c@{}}GGHL\cite{huang2022general}\end{tabular}}} &  \multicolumn{1}{c|}{{AutoAssign\cite{zhu2020autoassign}}} & \multicolumn{1}{c|}{{ \begin{tabular}[c]{@{}c@{}}75.34 \end{tabular}}} & {\begin{tabular}[c]{@{}c@{}}42.39\\ \end{tabular}}                          \\
						
						\multicolumn{1}{c|}{{\begin{tabular}[c]{@{}c@{}}GGHL\cite{huang2022general}\end{tabular}}} &  \multicolumn{1}{c|}{{GGHL\cite{huang2022general}}} & \multicolumn{1}{c|}{{ \begin{tabular}[c]{@{}c@{}}76.95 \end{tabular}}} & {\begin{tabular}[c]{@{}c@{}}42.39\\ \end{tabular}}                          \\

						\multicolumn{1}{c|}{{\begin{tabular}[c]{@{}c@{}}GGHL\cite{huang2022general}\end{tabular}}} &  \multicolumn{1}{c|}{{TAL+TAP of TOOD\cite{9710724})}} & \multicolumn{1}{c|}{{ \begin{tabular}[c]{@{}c@{}}77.11 \end{tabular}}} & {\begin{tabular}[c]{@{}c@{}}42.39\\ \end{tabular}}     
						
						                     \\ \hline
						\multicolumn{1}{c|}{{\begin{tabular}[c]{@{}c@{}}GGHL\cite{huang2022general}\end{tabular}}} &  \multicolumn{1}{c|}{{DTLA}} & \multicolumn{1}{c|}{{ \begin{tabular}[c]{@{}c@{}}\textbf{77.95} \end{tabular}}} & {\begin{tabular}[c]{@{}c@{}}42.39\\ \end{tabular}}                          \\ \hline
						
						\multicolumn{1}{c|}{{\begin{tabular}[c]{@{}c@{}}LO-Det\cite{huang2021lo}\end{tabular}}} &  \multicolumn{1}{c|}{{Anchor}} & \multicolumn{1}{c|}{{ \begin{tabular}[c]{@{}c@{}}66.17 \end{tabular}}} & {\begin{tabular}[c]{@{}c@{}}60.01 \\ \end{tabular}}                          \\
						
						\multicolumn{1}{c|}{{\begin{tabular}[c]{@{}c@{}}LO-Det\cite{huang2021lo}\end{tabular}}} &  \multicolumn{1}{c|}{{Centerness\cite{tian2019fcos}}} & \multicolumn{1}{c|}{{ \begin{tabular}[c]{@{}c@{}}69.65 \end{tabular}}} & {\begin{tabular}[c]{@{}c@{}}62.07\\ \end{tabular}}                          \\
												
						\multicolumn{1}{c|}{{\begin{tabular}[c]{@{}c@{}}LO-Det\cite{huang2021lo}\end{tabular}}} &  \multicolumn{1}{c|}{{ATSS\cite{zhang2020bridging}}} & \multicolumn{1}{c|}{{ \begin{tabular}[c]{@{}c@{}}70.01 \end{tabular}}} & {\begin{tabular}[c]{@{}c@{}}60.03\\ \end{tabular}}                          \\
																																	
						\multicolumn{1}{c|}{{\begin{tabular}[c]{@{}c@{}}LO-Det\cite{huang2021lo}\end{tabular}}} &  \multicolumn{1}{c|}{{AutoAssign\cite{zhu2020autoassign}}} & \multicolumn{1}{c|}{{ \begin{tabular}[c]{@{}c@{}}70.74 \end{tabular}}} & {\begin{tabular}[c]{@{}c@{}}62.07\\ \end{tabular}}                          \\

						\multicolumn{1}{c|}{{\begin{tabular}[c]{@{}c@{}}LO-Det\cite{huang2021lo}\end{tabular}}} &  \multicolumn{1}{c|}{{GGHL\cite{huang2022general}}} & \multicolumn{1}{c|}{{ \begin{tabular}[c]{@{}c@{}}71.26 \end{tabular}}} & {\begin{tabular}[c]{@{}c@{}}62.07\\ \end{tabular}}                          \\

						\multicolumn{1}{c|}{{\begin{tabular}[c]{@{}c@{}}LO-Det\cite{huang2021lo}\end{tabular}}} &  \multicolumn{1}{c|}{{TAL+TAP of TOOD\cite{9710724}}} & \multicolumn{1}{c|}{{ \begin{tabular}[c]{@{}c@{}}71.08 \end{tabular}}} & {\begin{tabular}[c]{@{}c@{}}62.07\\ \end{tabular}}                          \\
						
						\hline 
						
						\multicolumn{1}{c|}{{\begin{tabular}[c]{@{}c@{}}LO-Det\cite{huang2021lo}\end{tabular}}} &  \multicolumn{1}{c|}{{DTLA}} & \multicolumn{1}{c|}{{ \begin{tabular}[c]{@{}c@{}}\textbf{73.36} \end{tabular}}} & {\begin{tabular}[c]{@{}c@{}}62.07\\ \end{tabular}}                          \\

						\hline\hline
					\end{tabular}\vspace{0.5em}
				\end{threeparttable}
		}}
	}\vspace{-0.5em}
\end{table}

Finally, the TS-Conv model combining the above components is obtained. Compared with the GGHL model, the mAP$_{50}$, mAP$_{75}$ and mAP$_{50:95}$ of TS-Conv are improved by 2.34\% (+1.80), 5.45\% (+2.41), and 4.47\% (+1.98). The improvement in mAP$_{75}$ and mAP$_{50:95}$ metrics representing higher detection quality is more pronounced. In general, the ablation experiments validate the effectiveness of each component in the proposed TS-Conv. In addition, Table~\ref{table:3} lists the experimental results of different hyperparameter settings. The hyperparameter $\vartheta$ is used to measure the weight of statically pre-allocated labels versus dynamically allocated labels. The larger its value is, the more dependent it is on the label assignment of GGHL, which conforms to the Gaussian prior; the smaller its value is, the greater its reliance on the CNN adaptive learning score for label assignment. According to Table~\ref{table:3} we set $T=0.3$ and $\vartheta=0.5$ in TS-Conv to obtain the best results.

\begin{figure}[!tp]
	\centering
	\epsfig{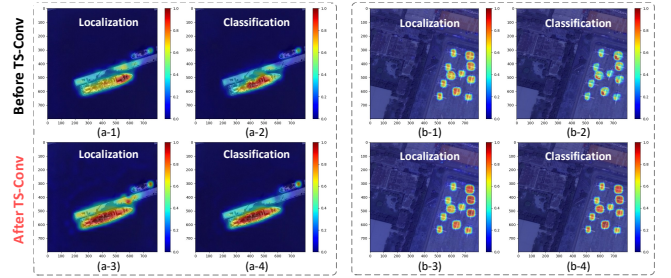}
	\caption{Visualization of feature sensitivity regions. The closer the color is to red, the higher the sensitivity of the feature at this location. The figures in rows 1 and 3 (Figs. (a-1), (a-2), (b-1) and (b-2)) are the visualization results of feature sensitivity regions before using the designed TS-Conv. The figures in rows 2 and 4 (Figs. (a-3), (a-4), (b-3) and (b-4)) are the visualization results of feature sensitivity regions after using the designed TS-Conv.}\label{fig:8}
\end{figure}
\begin{figure}[!tp]
	\centering
	\epsfig{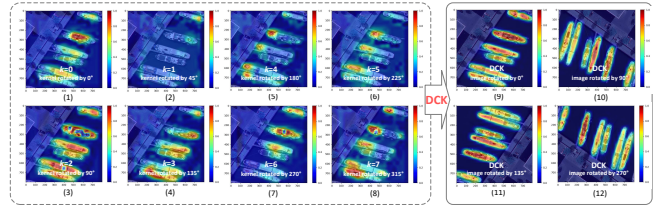}
	\caption{Visualization of features extracted by Dynamic circular kernel (DCK). Figs. (1)-(8) show the features extracted by convolutional kernels with different rotations in DCK, respectively. Figs. (9)-(12) show the DCK-extracted features of input images with different rotations, respectively.}\label{fig:9}
\end{figure}

\begin{figure}[!t]
	\centering
	\epsfig{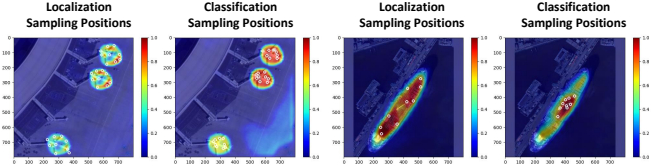}
	\caption{Visualization results of task-wise sampling positions.}\label{fig:R11}
\end{figure}

\textbf{2) Analysis of the IFS problem.} To more intuitively analyze the IFS problem faced by existing AOOD models like GGHL \cite{huang2022general}, Fig.~\ref{fig:8} visualizes the feature-sensitive regions of the localization and classification tasks. Fig.~\ref{fig:8} (a-1), (a-2) and (b-1), (b-2) show the feature sensitivity regions before using the proposed TS-Conv. The closer the color of the heatmap to red indicates that the model is more sensitive to the features in that region. From observing the feature sensitivity regions for various scenes and objects, it is obvious that the feature-sensitive regions of localization and classification tasks are significantly different. It is difficult for the existing schemes, such as decoupled-head used by CFC-Net \cite{ming2021cfc}, GGHL \cite{huang2022general}, etc., and shared-offset DCNs used by S$^2$ANet \cite{han2021align}, Oriented RepPoints \cite{li2022oriented}, etc., to take into account the respectively most sensitive features for different tasks. Correspondingly, Fig.~\ref{fig:8} (a-3), (a-4), (b-3) and (b-4) show the results of feature sensitivity regions after using the proposed TS-Conv. Sensitive features of different tasks located at different locations before input to TS-Conv are extracted and mapped to the same spatial location. The sensitive features of localization and classification are spatially aligned after TS-Conv's mapping. Then the optimal candidates are found among these spatially aligned features to predict the objects. It further demonstrates the effectiveness of the designed TS-Conv. Furthermore, Fig.~\ref{fig:R11} visualizes the sampling positions (white circles in the figure) of the model for localization and classification subtasks after using the designed sampling strategies. It can be seen from the visualization results that the feature sampling positions extract features from the corresponding task-sensitive regions, which shows the effectiveness of the task-wise sampling strategies in extracting more appropriate features. Besides, the localization feature sampling positions under explicit constraints also match the sensitive areas, indicating that the designed explicit constraints can effectively associate the feature-sensitive positions with the representation of OBB. In addition, it is found that for large-scale objects, such as the ship in Fig.~\ref{fig:R11}, the range of explicitly constrained sampling positions may be smaller than the object OBB during testing (not directly supervised by the ground truth of OBB during training), although these sampling positions still fit the shape and sensitive region of the OBB. This may be due to insufficient receptive field of the localization branch, which needs further study in the future.

\textbf{3) Analysis of orientation-robust features extracted by DCK.} Figs.~\ref{fig:9} (1)-(8) show the feature-sensitive regions of different oriented convolutional kernels in DCK (see details in Fig.~\ref{fig:5}), respectively. The different feature-sensitive regions indicate that the feature extraction is not robust to arbitrary orientations. Since objects' orientations in AOOD are more diverse than those in the ordinary OD, this problem further constrains the existing AOOD models' performance. Figs.~\ref{fig:9} (9) shows the feature-sensitive regions of the designed DCK, which adjust the optimal orientation and weights of eight-oriented kernels according to different inputs adaptively. Furthermore, Figs.~\ref{fig:9} (9)-(12) show the feature-sensitive regions of DCK when the input images are rotated in different orientations. The results indicate that the feature-sensitive regions of DCK are robust to arbitrary-oriented inputs, which do not fluctuate significantly with the objects' rotations. In addition, the performance of using the random rotation data augmentation (RRDA) strategy and DCK are compared in Table~\ref{table:3}. The performance of the model using only DCK exceeds that of the model using only the RRDA strategy. As shown in Table~\ref{table:3}, when using both the RRDA strategy and DCK, the mAP$_{50}$ is 0.28 higher than that of using DCK only and 0.83 higher than that of using RRDA only. The performance further confirms the designed DCK's ability to enhance the model's robustness to extract features with arbitrary orientations.

\subsection{Experiments for the Scalability of TS-Conv}
\begin{table}[tp]
	\centering
	\renewcommand\arraystretch{1.2}
	\setlength{\tabcolsep}{1mm}{
		\caption{\label{table:4}
			{Performance evaluation of the lightweight model using DTLA and TS-Conv-based knowledge distillation on the DOTAv1.0 dataset}}
		\resizebox{0.48\textwidth}{!}{\setlength{\tabcolsep}{0.8mm}{
				\begin{threeparttable}
					\begin{tabular}{c|ccccccc}
						\hline\hline
						\multirow{2}{*}{{Modules}} & \multirow{2}{*}{{mAP$_{50}$}}& \multirow{2}{*}{{\begin{tabular}[c]{@{}c@{}}Speed1\\ (fps)\end{tabular}}} & \multirow{2}{*}{{\begin{tabular}[c]{@{}c@{}}Speed2\\ (fps)\end{tabular}}} & \multirow{2}{*}{{\begin{tabular}[c]{@{}c@{}}Speed3\\ (fps)\end{tabular}}} & \multirow{2}{*}{{\begin{tabular}[c]{@{}c@{}}Speed4\\ (fps)\end{tabular}}}   & \multirow{2}{*}{{\begin{tabular}[c]{@{}c@{}}FLOPs\\ (G)\end{tabular}}} & \multirow{2}{*}{{\begin{tabular}[c]{@{}c@{}} Parameters\\  (MB)\end{tabular}}} \\ 
						{}& {} & {} & {}  & {} & {} & { }& { } \\
						\hline
						
						YOLOXnano \cite{ge2021yolox} & 64.26& 46.70 & 6.95& 22.50 & 3.59 & 15.87 & 0.91\\ \hline

						NanoDet-M \cite{ge2021yolox} & 63.84& 52.87 &7.51 & 23.24& 3.49& 15.11 & 0.95\\ 
						
						\hline
						YOLOv6nano \cite{ge2021yolox} & 70.82 & 49.53 & 6.40& 19.72 & 3.01& 15.87 & 4.30\\ \hline
						
						{LO-Det \cite{huang2021lo} (Baseline)} & {66.17}& {60.01} & {6.99}& {22.12} & 3.71 & {6.42} & {6.93}\\ \hline

						\begin{tabular}[c]{@{}c@{}} LO-Det \cite{huang2021lo} + GGHL \cite{huang2022general} \end{tabular}
						& \begin{tabular}[c]{@{}c@{}}71.26\\ \tiny{(+5.09)}\end{tabular} & 	\begin{tabular}[c]{@{}c@{}}62.07\\ \tiny{(+2.06)}\end{tabular} & 	\begin{tabular}[c]{@{}c@{}}7.68\\ \tiny{(+0.699)}\end{tabular}  & 	\begin{tabular}[c]{@{}c@{}}23.72\\ \tiny{(+1.60)}\end{tabular} & 	\begin{tabular}[c]{@{}c@{}}4.04\\ \tiny{(+0.33)}\end{tabular} & 	\begin{tabular}[c]{@{}c@{}}6.30\\ \tiny{(-0.12)}\end{tabular} & 	\begin{tabular}[c]{@{}c@{}}6.72\\ \tiny{(-0.21)}\end{tabular}   \\ \hline
						
						\begin{tabular}[c]{@{}c@{}}LO-Det \cite{huang2021lo} + DTLA \end{tabular}
						& \begin{tabular}[c]{@{}c@{}}73.36\\ \tiny{(+7.19)}\end{tabular} & 	\begin{tabular}[c]{@{}c@{}}62.07\\ \tiny{(+2.06)}\end{tabular} & 	\begin{tabular}[c]{@{}c@{}}7.68\\ \tiny{(+0.699)}\end{tabular}  & 	\begin{tabular}[c]{@{}c@{}}23.72\\ \tiny{(+1.60)}\end{tabular} & 	\begin{tabular}[c]{@{}c@{}}4.04\\ \tiny{(+0.33)}\end{tabular} & 	\begin{tabular}[c]{@{}c@{}}6.30\\ \tiny{(-0.12)}\end{tabular} & 	\begin{tabular}[c]{@{}c@{}}6.72\\ \tiny{(-0.21)}\end{tabular}   \\ \hline
						
						\begin{tabular}[c]{@{}c@{}} TS-Conv Lite \end{tabular}
						& \begin{tabular}[c]{@{}c@{}} \textbf{73.96}\\ \textbf{\tiny{(+7.79)}}\end{tabular} & 	\begin{tabular}[c]{@{}c@{}}62.07\\ \tiny{(+2.06)}\end{tabular} & 	\begin{tabular}[c]{@{}c@{}}7.68\\ \tiny{(+0.699)}\end{tabular}  & 	\begin{tabular}[c]{@{}c@{}}23.72\\ \tiny{(+1.60)}\end{tabular} & 	\begin{tabular}[c]{@{}c@{}}4.04\\ \tiny{(+0.33)}\end{tabular} & 	\begin{tabular}[c]{@{}c@{}}6.30\\ \tiny{(-0.12)}\end{tabular} & 	\begin{tabular}[c]{@{}c@{}}6.72\\ \tiny{(-0.21)}\end{tabular}   \\
						
						\hline\hline
					\end{tabular}
	\end{threeparttable}}}}\vspace{0.5em}
	\justifying{Note: The unit G is Giga, which represents $1\times10^{9}$. The unit MB represents $1\times10^{6}$ bytes. Speed1, Speed2, Speed3 and Speed4 are the detection speed on the RTX 3090 GPU, NVIDIA Jetson TX2, Jetson AGX Xavier, and Jetson Nano, respectively. The inference speed only includes the network inference speed without post-processing. TS-Conv Lite: TS-Conv Distilled LO-Det \cite{huang2021lo} + DTLA.}
\end{table}
\begin{table}[tbp]
	\centering
	\renewcommand\arraystretch{1.1}
	\setlength{\tabcolsep}{1mm}{
		\caption{\label{table:5}
			{Performance evaluation of the proposed TS-Conv for multimodal images on the DroneVehicle dataset}}
		\resizebox{0.48\textwidth}{!}{\setlength{\tabcolsep}{0.8mm}{
				\begin{threeparttable}
					\begin{tabular}{c|c|c|cccccc}
						\hline\hline
						Methods & RGB & Infrared & car & freight car & truck & bus & van & mAP$_{50}$ \\
						\hline 
						\multirow{2}{*}{RetinaNet (OBB) \cite{linFocalLossDense2017}} &   $\checkmark$ &   & 67.50 & 13.72 & 28.24 & 62.05 & 19.26 & 38.16\\
						~ & & $\checkmark$& 79.86 & 28.05 & 32.84 & 67.32 & 16.44 & 44.90 \\
						\cdashline{1-9}[0.8pt/2pt]
						\multirow{2}{*}{Faster R-CNN (OBB) \cite{xiaDOTALargeScaleDataset2018}} &  $\checkmark$ &  & 67.88 & 26.31 & 38.59 & 66.98 & 23.20 & 44.59\\
						~ & & $\checkmark$& 88.63 & 35.16 & 42.51 & 77.92 & 28.52 & 54.55 \\
						\cdashline{1-9}[0.8pt/2pt]
						\multirow{2}{*}{Faster R-CNN (Dpool) \cite{xiaDOTALargeScaleDataset2018}} &  $\checkmark$ &   & 68.23 & 26.40 & 38.73 & 69.08 & 26.38 & 45.76 \\
						~ & & $\checkmark$& 88.94 & 36.79 & 47.91 & 78.28 & 32.79 & 56.94 \\
						\cdashline{1-9}[0.8pt/2pt]
						\multirow{2}{*}{Mask R-CNN \cite{xiaDOTALargeScaleDataset2018}} &  $\checkmark$ &   & 68.23 & 26.40 & 38.73 & 69.08 & 26.38 & 45.76 \\
						~ & & $\checkmark$& 88.94 & 36.79 & 47.91 & 78.28 & 32.79 & 56.94 \\
						\cdashline{1-9}[0.8pt/2pt]
						
						\multirow{2}{*}{Cascade Mask R-CNN \cite{xiaDOTALargeScaleDataset2018}} &  $\checkmark$ &   & 68.00 & 27.25 & 44.67 & 69.34 & 29.80 & 47.81 \\
						~ & & $\checkmark$& 81.00 & 38.97 & 47.18 & 79.32 & 33.00 & 56.96 \\
						\cdashline{1-9}[0.8pt/2pt]
						
						\multirow{2}{*}{Hybrid Task Cascade \cite{xiaDOTALargeScaleDataset2018}} &  $\checkmark$ &   & 67.89 &27.22 &44.55 &70.22 &28.61 &47.70  \\
						~ & & $\checkmark$& 88.57 & 42.85 &47.71 &79.46 &34.16 &58.55 \\
						\cdashline{1-9}[0.8pt/2pt]
						
						\multirow{2}{*}{RoITransformer \cite{ding2019learning}} &  $\checkmark$ &   & 68.13& 29.08& 44.17& 70.55& 27.64 &47.91  \\
						~ & & $\checkmark$& 88.85& 41.49& 51.53& 79.48& 34.39& 59.15 \\
						\cdashline{1-9}[0.8pt/2pt]
						
						\multirow{2}{*}{ReDet \cite{han2021redet}]} &  $\checkmark$ &   & 69.48 &31.46 &47.87 &77.37 &29.03 &51.04  \\
						~ & & $\checkmark$& 89.47 &42.82 &53.95 &79.89 &36.56 &60.54 \\
						\cdashline{1-9}[0.8pt/2pt]
						
						\multirow{2}{*}{Gliding Vertex \cite{xu2020gliding}} &  $\checkmark$ &   & 75.77& 33.75& 46.08& 68.05& 38.72& 52.48  \\
						~ & & $\checkmark$& 89.15 &42.95 &59.72 &78.75 &43.88 &62.89 \\
						\cdashline{1-9}[0.8pt/2pt]
						
						\multirow{2}{*}{GGHL \cite{huang2022general}} &  $\checkmark$ &   & 89.95 & 47.19 & 65.67 & 91.66 & 44.84 & 67.86  \\
						~ & & $\checkmark$& 94.32 & 53.45 & 64.63 & 90.85 & 51.38 & 70.93 \\
						\cdashline{1-9}[0.8pt/2pt]
						
						UA-CMDet \cite{sun2022drone} &  $\checkmark$ & $\checkmark$  & 87.51 &46.80 &60.70 &87.08 &37.95 & 64.01  \\
						\cdashline{1-9}[0.8pt/2pt]
						
						\hline
						\textbf{TS-Conv}&  $\checkmark$ &  &  90.07 & 48.12 & 64.39 & 91.67 & 46.52 & 68.15 \\
						\textbf{TS-Conv} &   & $\checkmark$ &  94.55 & 53.70 & 64.91& 91.15 & 52.02 & 71.27 \\
						\textbf{TS-Conv} &  $\checkmark$ & $\checkmark$ &  \textbf{94.87} & \textbf{55.16} & \textbf{65.93} & \textbf{92.04} & \textbf{53.64} & \textbf{72.33} \\
						\cdashline{1-9}[0.8pt/2pt]
						
						\hline\hline
					\end{tabular}
	\end{threeparttable}}}}\vspace{0.5em}
	\justifying{Note: The testing image size is 640 $\times$640 pixels. The TS-Conv$^*$ represents that it not only samples the features of different tasks separately, but also samples the features of different modalities.}
\end{table}

\textbf{1) Scalability of TS-Conv on lightweight models.} Although TS-Conv improves detection performance and robustness, the introduction of DCNs also brings additional computational burdens compared to the baseline model GGHL \cite{huang2022general}. For the benefits of TS-Conv to be applied in the lightweight model, the experiments on different embedded devices are designed and validated, as listed in Table~\ref{table:4}. The lightweight AOOD model LO-Det \cite{huang2021lo} is chosen as the baseline. First, the anchor-based label assignment strategy of LO-Det is improved to GGHL and the proposed DTLA for comparison. The results in Table~\ref{table:4} demonstrate that using the proposed dynamic label assignment strategy DTLA further improves the performance of the lightweight model by 2.95\% (+2.10) without losing model inference efficiency and increasing model complexity compared to using the static label assignment strategy GGHL. Second, the knowledge distillation is adopted to make DTLA take full advantage of the task-wise sensitive features learned by TS-Conv without complicating the lightweight model structure. Specifically, The LO-Det+DTLA is utilized as the student model, and the TS-Conv is used as the teacher model. RGB images and IR images are used to train one teacher model each, and then two teacher models are used to distill knowledge to the student model according to the DKED strategy \cite{huang2022extracting}. The model obtained from this scheme is denoted as TS-Conv Lite. The results show that Ts-Conv Lite's performance is further improved without additional inference cost. TS-Conv Lite's mAP$_{50}$ improves 11.77\% (+7.79) compared to the baseline model with fewer model parameters and faster inference because it does not rely on anchor boxes. In addition, experiments also compare the performance of TS-Conv Lite and other lightweight models. The results show that the proposed TS-Conv Lite has better performance and faster detection speed on embedded devices.

\begin{figure}[!t]
	\centering
	\epsfig{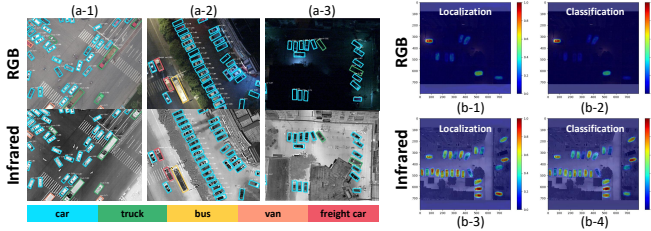}
	\caption{Visualization results of TS-Conv on the DroneVehicle dataset. Figs. (a-1)-(a-3) show the object detection results of RGB-Infrared multimodal image pairs under different lighting conditions. Figs. (b-1)-(b-4) show the feature sensitivity regions of different modalities and different tasks for RGB-Infrared multimodal image pairs.}\label{fig:10}
\end{figure}

\textbf{2) Scalability of TS-Conv for multimodal data.} The effectiveness of the proposed TS-Conv for multimodal data is evaluated on the multimodal AOOD dataset DroneVehicle \cite{sun2022drone}. The results are shown in Table~\ref{table:5} and Fig.~\ref{fig:10}. The performance of the proposed TS-Conv and the baseline model GGHL are evaluated on RGB images, infrared images, and RGB-infrared image pairs, respectively. The results demonstrate that TS-Conv has improved performance compared with GGHL in all three groups of experiments. It illustrates the effectiveness of TS-Conv on multimodal data and demonstrates the merits of TS-Conv for improving the accuracy of oriented bounding boxes again.
\begin{table*}[tp]
	\scriptsize
	\centering
	\renewcommand\arraystretch{1}
	\setlength{\tabcolsep}{12mm}{
		\caption{\label{table:6}
			{Comparative experiments on the DOTAv1.0 dataset}}
		\resizebox{\textwidth}{!}{\setlength{\tabcolsep}{0.55mm}{
				\begin{threeparttable}
					\centering
					\begin{tabularx}{\textwidth}{c|c|c|c|ccccccccccccccc|c|c}
						\hline\hline
						\multirow{2}{*}{Methods}                                  & \multirow{2}{*}{Backbone} & \multirow{2}{*}{Stage}& \multirow{2}{*}{Anchor} & \multirow{2}{*}{PL} & \multirow{2}{*}{BD} & \multirow{2}{*}{BR} & \multirow{2}{*}{GTF} & \multirow{2}{*}{SV} & \multirow{2}{*}{LV} & \multirow{2}{*}{SH} & \multirow{2}{*}{TC} & \multirow{2}{*}{BC} & \multirow{2}{*}{ST} & \multirow{2}{*}{SBF} & \multirow{2}{*}{RA} & \multirow{2}{*}{HA} & \multirow{2}{*}{SP} & \multirow{2}{*}{HC} & \multirow{2}{*}{mAP$_{50}$} & \multirow{2}{*}{\begin{tabular}[c]{@{}c@{}}Speed\\ (fps)\end{tabular}} \\
						& &  &  &  &  &  &  & &  &  & &  &  & & & & & &  &  \\ \hline
						ROI Trans. \cite{ding2019learning} & R-101 & Two & AB & 88.53& 77.91 & 37.63& 74.08 & 66.53& 62.97  & 66.57& 90.50& 79.46& 76.75& 59.04  & 56.73& 62.54  & 61.29 & 55.56& 67.74& 7.80 \\
						SCRDet \cite{yangSCRDetMoreRobust2019} & R-101 & Two & AB & 89.98 & 80.65 & 52.09& 68.36 & 68.36 & 60.32 & 72.41 & 90.85& 87.94& 86.86 & 65.02 & 66.68  & 66.25  & 68.24  & 65.21  & 72.61  & 9.51\\
						RSDet \cite{qian2019learning} & R-101 & Two & AB & 89.80 & 82.90  & 48.60& 65.20& 69.50 & 70.10 & 70.20 & 90.50  & 85.60 & 83.40 & 62.50 & 63.90& 65.60 & 67.20& 68.00  & 72.20 & -  \\	
						Gliding Vertex \cite{xu2020gliding} & R-101 & Two & AB & 89.64 & 85.00 & 52.26 & 77.34 & 73.01 & 73.14 & 86.82 & 90.74& 79.02  & 86.81 & 59.55  & 70.91 & 72.94 & 70.86 & 57.32 & 75.02 & 13.10 \\			
						CSL \cite{yang2020arbitrary}& R-152 & Two & AB & \textbf{90.25}  & 85.53 & 54.64  & 75.31 & 70.44  & 73.51 & 77.62 & 90.84 & 86.15 & 86.69  & 69.60 & 68.04 & 73.83 & 71.10 & 68.93 & 76.17 & 8.89 \\			
						Oriented R-CNN$*$\cite{xie2021oriented} & R-50 & Two & AB & 89.84 & 85.43&  \textbf{61.09} & 79.82 & 79.71& 85.35 & 88.82 & 90.88&  86.68 & 87.73 & 72.21 & 70.80 & \textbf{82.42} & 78.18&  74.11 & 80.87
						& 8.10\\
												
						CGCDet \cite{10042179}& R-50 &Two& AB & 88.93 & 84.45 & 53.93 & 78.56 & 78.54 & 82.46 & 87.90 & 90.87 & 87.46 & 84.79 & 65.56 & 63.45 &76.15 & 71.58 &65.32 & 77.34 & 7.80 \\
						CGCDet$*$\cite{10042179}  & R-50 & Two& AB & 89.42 & 84.49 & 59.83 & 80.78 & 79.53 & 84.75 & 88.55 & 90.79 & 87.81 & 87.06 & 69.72 & 71.09 & 79.38 & 80.96 & 75.32 & 80.70 & $<2$ \\
						DODet \cite{9706434} & R-50 & Two & AB &  {89.34} & {84.31} & {51.39} & {71.04} & {79.04} & {82.86} & {88.15} & {90.90} & {86.88} & {84.91} & {62.69} & {67.63} & {75.47} & {72.
							22} & {45.54} & {75.49} & {7.98} \\
						{DODet$*$\cite{9706434}} & { R-50} & { Two} & { AB} & {  89.96} & { 85.52} & { 58.01} & { 81.22} & { 78.71} & { 85.46} & { 88.59} & { 90.89} & { 87.12} & { 87.80} & { 70.50} & { 71.54} & { 82.06} & { 77.43} & { 74.47} & {80.62} & { 3.41} \\
						
						\hline
						O$^2$-DNet \cite{wei2020oriented} & H-104 & One & AF & 89.31 & 82.14  & 47.33  & 61.21  & 71.32 & 74.03 & 78.62& 90.76 & 82.23 & 81.36  & 60.93 & 60.17 & 58.21 & 66.98 & 61.03 & 71.04 & - \\
						BBAVectors \cite{yi2020oriented} & R-101& One & AF & 88.35& 79.96& 50.69 & 62.18& 78.43& 78.98 & 87.94 & 90.85 & 83.58& 84.35& 54.13& 60.24 & 65.22 & 64.28 & 55.70 & 72.32 & 18.37 \\
						CFC-Net \cite{ming2021cfc}& R-50  & One & AB & 89.08 & 80.41 & 52.41 & 70.02 & 76.28 & 78.11 & 87.21 & 90.89 & 84.47  & 85.64 & 60.51 & 61.52 & 67.82 & 68.02 & 50.09 & 73.50 & 17.81 \\
						RIDet \cite{ming2021optimization} & R-101 & One & AB & 88.94 & 78.45 & 46.87 & 72.63 & 77.63 & 80.68 & 88.18 & 90.55 & 81.33 & 83.61 & 64.85 & 63.72 & 73.09 & 73.13 & 56.87 & 74.70 & 13.36\\
						GGHL \cite{huang2022general} & D-53 & One & AF & 89.74 & 85.63  & 44.50  & 77.48 & 76.72  & 80.45 & 86.16 & 90.83 & \textbf{88.18} & 86.25 & 67.07 & 69.40 & 73.38 & 68.45 & 70.14 & 76.95 & 42.30 \\					
						PolarDet \cite{zhao2021polardet} & R-101 & One & AF & 89.65& 87.07 &48.14 &70.97 &78.53 &80.34& 87.45& 90.76 &85.63 &86.87 &61.64 &70.32& 71.92& 73.09& 67.15& 76.64 & 25.00 \\
						GWD$*$\cite{yang2021rethinking} & R-152 & One & AB & 86.96 & 83.88 & 54.36 & 77.53 & 74.41 & 68.48 & 80.34 & 86.62 & 83.41 & 85.55 & \textbf{73.47} & 67.77 & 72.57 & 75.76 & 73.40 & 76.30 & 13.86 \\
						KFIoU$*$\cite{yang2022kfiou} & R-152 & One & AB &89.46 &85.72 &54.94 &80.37 &77.16 &69.23 &80.90 &90.79 &87.79 &86.13 &73.32 &68.11 & 75.23 & 71.61 & 69.49 & 77.35 & 13.79 \\ \hline
						R$^3$Det \cite{yang2019r3det} & R-152 & Refine & AB & 89.80 & 83.77 & 48.11 & 66.77 & 78.76 & 83.27 & 87.84 & 90.82& 85.38 & 85.51& 65.67 & 62.68 & 67.53 & 78.56 & 72.62 & 76.47  & 12.39 \\
						\cdashline{1-21}[0.8pt/2pt]	 
						\multirow{2}{*}{S$^2$A-Net$*$ \cite{han2021align}} & R-50 & Refine & AB  & 89.07  & 82.22 & 53.63  & 69.88 & 80.94  & 82.12 & 88.72  & 90.73 & 83.77 & 86.92 & 63.78 & 67.86 & 76.51 & 73.03& 56.60 & 76.38 & 17.60\\
						~ & R-101 & Refine & AB & 88.89 & 83.60 & 57.74 & \textbf{81.95} & 79.94 & 83.19  & \textbf{89.11} & 90.78& 84.87 & \textbf{87.81} & 70.30 & 68.25 & 78.30 & 77.01 & 69.58 & 79.42 & 13.79\\
						\cdashline{1-21}[0.8pt/2pt]
						\multirow{2}{*}{G-Rep$*$ \cite{hou2022g}} & R-50 & Refine & AF & 87.76 &81.29 &52.64 &70.53 &80.34 &80.56 &87.47 &90.74 &82.91 &85.01 &61.48& 68.51 &67.53 &73.02 &63.54 &75.56 & 14.74 \\
						~ & RX-101 & Refine & AF & 88.98 &79.21 &57.57 &74.35 &81.30 &85.23 &88.30 &90.69& 85.38 &85.25 &63.65 &68.82 &77.87 &78.76& 71.74 & 78.47 & - \\
						\cdashline{1-21}[0.8pt/2pt]
						\multirow{3}{*}{\begin{tabular}[c]{@{}c@{}}Oriented \\ RepPoints$*$ \cite{li2022oriented}\end{tabular}} & R-50 & Refine & AF & 87.02 & 83.17  & 54.13  & 71.16 & 80.18  & 78.40 & 87.28 & \textbf{90.90} & 85.97 & 86.25 & 59.90 & 70.49 & 73.53 & 72.27 & 58.97 & 75.97 & 16.10 \\
						~ & R-101 & Refine & AF & 89.53 & 84.07  & 59.84 & 71.76 & 79.95 & 80.03 & 87.33 & 90.84 & 87.54 & 85.23 & 59.15 & 66.37 & 75.23 & 73.75 & 57.23 & 76.52 & 14.23 \\
						~ & Swin-T & Refine & AF & 88.72 & 80.56  & 55.69  & 75.07 & \textbf{81.84}  & 82.40 & 87.97 & 90.80 & 84.33 & 87.64 & 62.80 & 67.91 & 77.69 & \textbf{82.94} & 65.46 & 78.12 & -\\
						\cdashline{1-21}[0.8pt/2pt]

					\cdashline{1-21}[0.8pt/2pt]
						
						{R$^3$Det+GWD$*$ \cite{yang2021rethinking}} & { R-152} & { Refine} & { AB} & {  89.66} & { 84.99} & { 59.26} & { 82.19} & {78.97} & {84.83} & {87.70} & {90.21} & {86.54} & {86.85} & {73.47} & {67.77} & {76.92} & {79.22} & {74.92} & { 80.23} & { 12.40}\\
						{R$^3$Det+KLD$*$ \cite{yang2022kfiou} }& { R-152} & { Refine } & { AB} & { 89.92} & {85.13} & {59.19} & {81.33 } & { 78.82} & {84.38}& { 87.50}& { 89.80} & {87.33} & {87.00} & {72.57} & {71.35} & {77.12} & {79.34} & { \textbf{78.68}} & {80.63} & { 12.39}\\
{R$^3$Det+KFIoU$*$ \cite{yang2022kfiou}} & { R-152} & { Refine} & { AB} & { 88.89} & { 85.14 }& { 60.05} & { 81.13} & { 81.78} & { \textbf{85.71}} & { 88.27} & { 90.87 }& { 87.12} & { 87.91 }& { 69.77 }& { 73.70} & { 79.25 }& { 81.31 }& { 74.56} & { \textbf{81.03}} & { 9.10 }\\ 

{R$^3$Det+KFIoU$*$ \cite{yang2022kfiou}} & { Swin-T }& { Refine} & { AB} & { 89.50} & { 84.26} & { 59.90 }& { 81.06} & { 81.74} & { 85.45 }& { 88.77} & { 90.85}& { 87.03} & { 87.79} & { 70.68}& { \textbf{74.31}} & { 78.17}& { 81.67}& { 72.37} & {80.90} & { 8.85 }\\
\hline
\textbf{TS-Conv} & D-53 & Refine & AF & 89.86 & 87.05 & 49.12 & 74.01 & 78.97 & 81.28 & 88.24 & 90.77 & 86.85 & 87.24 & 71.87 & 69.88 & 77.01 & 70.43 & 78.63 & 78.75 & 23.23 \\
{\textbf{TS-Conv}$*$} & {D-53} & {Refine} & {AF} & {89.98} & {\textbf{88.24}} & {59.85} & {80.09} & {79.57} & {83.27} & {89.10} & {90.89} & {88.00} & {87.65} & {71.49} & {70.66} & {77.28} & {80.14} & {78.37} & {80.97} & {16.49} \\

\textbf{TS-Conv Lite} & Mobile2 & One & AF & 89.08 & 84.20 & 38.08 & 74.47 & 77.03 & 75.40 & 86.50 & 90.84 & 79.44 & 85.63 & 59.33 & 66.51 & 67.03 & 67.73 & 68.12 & 73.96 & \textbf{62.07} \\
						
						\hline\hline
					\end{tabularx}
	\end{threeparttable}}}}\vspace{0.5em}
	\justifying{Note: Bold font indicates the best results. The backbone networks R50, R-101, R-152, H-104, D-53, RX-101, Swin-T, and Mobile2 represent the ResNet50, ResNet101, ResNet152, Hourglass104, DarkNet53, ResNeXt101, Swin Transformer Tiny, and MobileNetv2, respectively. “AF” represents anchor-free methods, and 'AB' represents anchor-based methods. The inference speed only includes the network inference speed (batch size=1) on an RTX 3090 GPU. The speed of some methods could not be tested due to the available codes, which is indicated by “-”. “$*$” represents multi-scale training and multi-scale testing.} 
\end{table*}
TS-Conv has discussed the differences in task-wise feature-sensitive regions, furthermore, the differences in modality-wise feature-sensitive regions are analyzed here. Figs.\ref{fig:10} (b-1)-(b-4) show the differences between task-wise and modality-wise features for RGB-Infrared image pairs. RGB images contain richer color and texture features but are more susceptible to lighting conditions. Infrared images highlight objects in low light conditions but lack more detailed features. Therefore, TS-Conv is extended to sample modality-wise features separately along the lines of "separate sampling and aligned mapping". This model is denoted as TS-Conv$^*$. The results in Table~\ref{table:5} reflect the effectiveness of the modality-wise samplings strategy for further improving the detection performance on the multimodal dataset. In the future, this problem is expected to be explored in more depth. In addition, experiments with lightweight models are also conducted to verify the scalability of TS-Conv on multimodal data.

\subsection{Comparison Experiments}
In this subsection, the performance of the proposed TS-Conv and state-of-the-art methods is compared on several datasets, including DOTA \cite{xiaDOTALargeScaleDataset2018,li2020object}, DIOR-R \cite{cheng2022anchor}, HRSC2016 \cite{liu2017high}, SSDD+ \cite{li2017ship}, SKU-110KR \cite{pan2020dynamic}, etc.
\begin{table}[!t]
	\centering
	\renewcommand\arraystretch{1}
	\setlength{\tabcolsep}{1.3mm}{
		\caption{\label{table:7}
			{Comparative performance of different methods on DOTAv1.0, DOTAv1.5, and DOTAv2.0 datasets}}
		\resizebox{0.48\textwidth}{!}{\setlength{\tabcolsep}{1.2mm}{
				\begin{threeparttable}	
					\begin{tabular}{c|ccc}
						\hline\hline
						Methods   & 
						\begin{tabular}[c]{@{}c@{}}mAP$_{50}$@v1.0\end{tabular}
						& 						\begin{tabular}[c]{@{}c@{}}mAP$_{50}$@v1.5\end{tabular} & 						\begin{tabular}[c]{@{}c@{}}mAP$_{50}$@v2.0\end{tabular} \\ \hline
						RetinaNet OBB \cite{linFocalLossDense2017} & 66.28 & 59.16 & 46.68 \\
						Mask R-CNN \cite{ding2021object} & 70.71 & 62.67 & 49.47 \\
						Cascade Mask R-CNN \cite{ding2021object} & 70.96 & 63.41 & 50.04 \\
						Hybrid Task Mask \cite{ding2021object} & 71.21 & 63.40 & 50.34 \\
						Faster R-CNN OBB \cite{renFasterRCNNRealTime2017a} & 69.36 & 62.00 & 47.31 \\
						Faster R-CNN OBB + Dpool \cite{ding2021object} & 70.14 & 62.20 & 48.77 \\		
						Faster R-CNN H-OBB \cite{ding2021object} & 70.11 & 62.57 & 48.90 \\	
						Faster R-CNN OBB + RT \cite{ding2021object} & 73.76 & 65.03 & 52.81 \\						
						\hline
						GGHL \cite{huang2022general} (Baseline) & 73.98  & 68.92 & 57.17 \\ 
						\textbf{TS-Conv} & \textbf{75.04 \tiny{(+1.06)}} & \textbf{71.18 \tiny{(+2.86)}} & \textbf{59.77 \tiny{(+2.60)}}   \\
						\hline\hline
					\end{tabular}
	\end{threeparttable}}}}\vspace{0.5em}
	\justifying{Note: Bold font indicates the best results. In order to make a fair comparison with the methods in the DOTAv2.0 benchmark \cite{ding2021object}, the experiments above do not use data augmentation and other tricks like these comparison methods. $mAP_{50}$@v1.0, $mAP_{50}$@v1.5, and $mAP_{50}$@v2.0 denote the results on the DOTAv1.0, DOTAv1.5, and DOTAv2.0 datasets \cite{li2020object}, respectively.}
\end{table}

\textbf{1) Comparison experiments on the DOTA datasets.} Table~\ref{table:6} provides the performance comparison results of the different methods on the most widely used AOOD dataset DOTAv1.0. The experimental results show that the performance and speed of the proposed TS-Conv outperform most of the AOOD methods (single-scale: mAP$_{50}$=78.75, multi-scale: mAP$_{50}$=80.97), further validating the effectiveness of TS-Conv. Although the performance of TS-Conv (single-scale testing) is slightly lower than that of the two-stage method Oriented R-CNN \cite{xie2021oriented} and the refine-stage method S$^2$ANet (with the larger backbone ResNet-101) \cite{han2021align} when single-scale testing are used, the detection speed of TS-Conv is much faster than those of Oriented R-CNN \cite{xie2021oriented} and S$^2$ANet \cite{han2021align}. Furthermore, after using multi-scale training and testing, the proposed TS-Conv has better mAP performance than R$^3$Det+GWD (refine-stage), R$^3$Det+KLD (refine-stage), CGCDet (two-stage), DODet (two-stage) and other methods. Its mAP is only slightly lower than R$^3$Det+KFIoU (ResNet152) by 0.06\% but TS-Conv is significantly faster than these comparison methods. Besides, the proposed TS-Conv is an anchor-free method, which is more flexible and does not rely on many hyperparameters of anchor boxes. 
In addition, the performance of the lightweight model TS-Conv Lite (mAP$_{50}$=73.96) can reach the level of many larger models and has a detection speed that far exceeds that of other methods. Furthermore, the performance evaluation on the latest versions of the DOTA datasets, i.e., DOTAv1.5 and DOTAv2.0 \cite{li2020object} are listed in Table~\ref{table:7}. These datasets cover a wider category of objects and more small objects that are difficult to detect. The results also demonstrate the performance advantage of the proposed TS-Conv over existing methods.
\begin{table}[t]
	\centering
	\renewcommand\arraystretch{1.2}
	\setlength{\tabcolsep}{1mm}{
		\caption{\label{table:8}
			{Comparative performance of different methods on the HRSC2016 dataset}}
		\resizebox{0.48\textwidth}{!}{\setlength{\tabcolsep}{1mm}{
				\begin{tabular}{c|c|c|ccc}
					\hline\hline
					Method  & Anchor  & Backbone  & mAP$_{50}$(07) & mAP$_{50}$(12) & mAP$_{75}$(07) \\
					\hline
					R$^2$CNN \cite{jiang2017r2cnn}     & AB      & ResNet101 & 73.07   & 79.73   &-\\
					RoI-Transformer \cite{ding2019learning} & AB & ResNet101 & 86.20   & -      &- \\
					Gliding Vertex \cite{xu2020gliding}  & AB    & ResNet101 & 88.20   & -       &-\\
					BBAVectors \cite{yi2020oriented}  & AF     & ResNet101 & 88.60    & -       &-\\
					CenterMap OBB \cite{wang2020learning}  & AB  & ResNet50  & -       & 92.80   &- \\
					RetinaNet-R \cite{yang2019r3det}    & AB     & ResNet101 & 89.18   & 95.21  &- \\
					RetinaNet-GWD \cite{yang2021rethinking} & AB & ResNet50  & 85.56 & - & 60.31 \\ 
					RetinaNet-KLD \cite{yang2021learning} & AB & ResNet50  & 87.45 & - & 72.39 \\
					R$^3$Det \cite{yang2019r3det}     & AB       & ResNet101 & 89.26   & 96.01  &- \\
					R$^3$Det-DCL \cite{yang2021dense}   & AB     & ResNet101 & 89.46   & 96.41   &-\\
					R$^3$Det-GWD \cite{yang2021rethinking} & AB & ResNet50  & 89.43 & - & 68.88 \\
					R$^3$Det-KLD \cite{yang2021learning} & AB & ResNet50  & 89.97 & - & 77.38 \\
					S$^2$ANet \cite{han2021align}     & AB       & ResNet101 & 90.17   & 95.01   &-\\
					Oriented RepPoints \cite{li2022oriented} & AF & ResNet50  & 90.40   & 97.26   &- \\
					\hline
					GGHL \cite{huang2022general} (Baseline)    & AF   & DarkNet53 & 89.53  & 96.50 & 76.07 \\
					\textbf{TS-Conv}    & AF  & DarkNet53 & \textbf{90.59 \tiny{(+1.06)}}  & \textbf{97.64 \tiny{(+1.14)}}  &  \textbf{78.34 \tiny{(+2.27)}}\\
					\hline\hline
	\end{tabular}}}}\vspace{0.5em}
	\justifying{Note: Bold font indicates the best results. AF represents anchor-free methods, and AB represents anchor-based methods. The mAP$_{50}$(07) and mAP$_{50}$(12) represent the mAP calculated on standard of VOC07 and VOC12, respectively.}
\end{table}
\begin{table}[t]
	\centering
	\renewcommand\arraystretch{1.1}
	\setlength{\tabcolsep}{0.5mm}{
		\caption{\label{table:9}
			{Comparative performance of different methods on the DIOR-R dataset}}
		\resizebox{0.48\textwidth}{!}{\setlength{\tabcolsep}{1.2mm}{
				\begin{threeparttable}	
					\begin{tabular}{c|c|c|ccc}
						\hline\hline
						Methods &  Anchor & Backbone & mAP$_{50}$  & mAP$_{75}$    & mAP$_{50:95}$     \\
						\hline
						RetinaNet-O \cite{linFocalLossDense2017}  & AB & ResNet50  & 57.55   & -    & -     \\
						Faster RCNN-O \cite{renFasterRCNNRealTime2017a}  & AB  & ResNet50  & 59.54    & -  & -      \\
						Gliding Vertex \cite{xu2020gliding} & AB  & ResNet50  & 60.06   & -    & -     \\
						RoI-Transformer \cite{ding2019learning}  & AB    & ResNet50   & 63.87      & -  & -    \\
						AOPG \cite{cheng2022anchor}   & AF   & ResNet50 & 64.41     & -    & -   \\
						Oriented RepPoints \cite{li2022oriented} & AF & ResNet50 & 66.71     & -    & -   \\
						\hline
						GGHL \cite{huang2022general} (Baseline) & AF  & DarkNet53 & 66.48  & 36.99    & 37.44      \\
						\textbf{TS-Conv} & AF & DarkNet53 & \textbf{68.47 \tiny{(+1.99)}} & \textbf{42.69 \tiny{(+5.70)}}& \textbf{41.38 \tiny{(+3.94)}} \\
						\hline\hline
					\end{tabular}
	\end{threeparttable}}}}\vspace{0.5em}
	\justifying{Note: Bold font indicates the best results. AF represents anchor-free methods, and AB represents anchor-based methods.}
\end{table}
\begin{table}[t]
	\centering
	\renewcommand\arraystretch{1.1}
	\setlength{\tabcolsep}{1.2mm}{
		\caption{\label{table:10}
			{Comparative performance of different methods on the SSDD+ dataset}}
		\resizebox{0.48\textwidth}{!}{\setlength{\tabcolsep}{1.5mm}{
				\begin{threeparttable}
					\begin{tabular}{c|c|c|c|c|c}
						\hline\hline
						\multirow{2}{*}{Methods} & \multirow{2}{*}{Anchor} & \multirow{2}{*}{Backbone}& \multirow{2}{*}{mAP$_{30}$}  & \multirow{2}{*}{mAP$_{50}$} & \multirow{2}{*}{mAP$_{75}$} \\
						& &  &   & &  \\ \hline
						DRBox-v1 \cite{wang2018simultaneous}  &  AB  & VGG16  & 86.41 & - & -   \\
						SDOE \cite{wang2018simultaneous}  &  AB  & VGG16 & -  & 82.40  & -   \\
						DRBox-v2 \cite{an2019drbox} & AB  & VGG16  & 92.81 & 85.17 & - \\\hline
						\begin{tabular}[c]{@{}c@{}}GGHL \cite{huang2022general} \\(Baseline1) \end{tabular}& AF  &  DarkNet53 & 95.10 & 90.22 & 22.18 \\						\cdashline{1-6}[0.8pt/2pt]
						\textbf{TS-Conv} & AF  &  DarkNet53 & \textbf{96.56 \tiny{(+1.46)}} & \textbf{92.48 \tiny{(+2.26)}} & \textbf{41.96 \tiny{(+19.78)}}  \\\hline
						\begin{tabular}[c]{@{}c@{}} LO-Det + GGHL \cite{huang2021lo} \\ (Baseline2)\end{tabular} & AF  & MobileNetv2  & 93.87 & 85.90 & 16.64 \\						\cdashline{1-6}[0.8pt/2pt]
						\textbf{LO-Det\cite{huang2021lo} + DTLA} & AF & MobileNetv2  & \textbf{93.89 \tiny{(+0.02)}} & \textbf{87.08 \tiny{(+1.18)}} & \textbf{26.73 \tiny{(+10.09)}} \\ 
						\hline\hline
					\end{tabular}
	\end{threeparttable}}}}\vspace{0.5em}
	\justifying{Note: The testing image size is 800 $\times$800 pixels. To be consistent with the comparison method, the confidence threshold is set to 0.2. AF represents anchor-free methods, and AB represents anchor-based methods.}
\end{table}
\begin{figure}[!t]
	\centering
	\epsfig{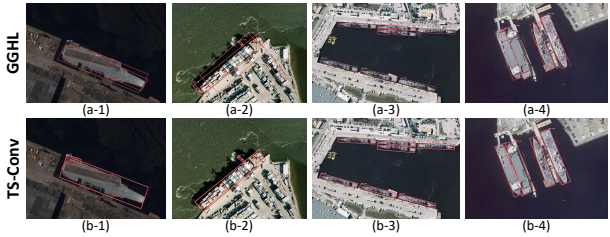}
	\caption{Comparison of visualization results between GGHL and TS-Conv on the HRSC2016 dataset. Figs. (a-1)-(a-4) show the detection results of GGHL, and Figs. (b-1)-(b-4) show the detection results of TS-Conv.}\label{fig:12}
\end{figure}

\begin{figure}[!ht]
	\centering
	\epsfig{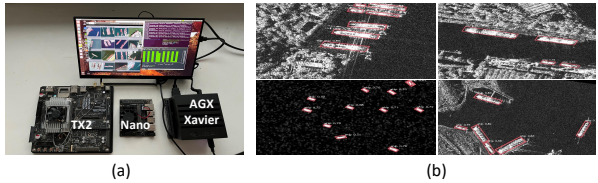}
	\caption{Presentation of other experiments. (a) Embedded edge devices, including Nvidia Jetson TX2, Nano and AGX Xavier, for evaluating the performance of lightweight models. (b) Visualization results of the proposed TS-Conv on the SSDD+ dataset.}\label{fig:14}
\end{figure}

\textbf{2) Comparison experiments on other datasets.} The results of comparison experiments on the HRSC2016 dataset are listed in Table~\ref{table:8}. TS-Conv also performs better than the existing methods in the ship detection, where the aspect ratio of objects is significant. Fig.~\ref{fig:12} shows that TS-Conv predicts more accurately and with fewer false alarms for oriented bounding boxes compared to GGHL \cite{huang2022general}. As shown in Table~\ref{table:9}, TS-Conv also outperforms the existing methods on the DIOR-R dataset \cite{cheng2022anchor}, where objects have more scale variations and categories. In particular, the improvements are more significant for mAP$_{75}$ and mAP$_{50:95}$. The results on the SAR dataset SSDD+ are given in Table~\ref{table:10}. The results demonstrate that TS-Conv performs better on the datasets with other data modality. More visualized results are shown in Fig.~\ref{fig:14}. In summary, extensive comparison experiments are conducted on datasets covering multiple scenes, multimodal images (RGB, infrared, SAR, and panchromatic images), multiple categories of objects, and different lighting conditions (daytime and nighttime). The state-of-the-art results demonstrate the effectiveness and generality of TS-Conv.

%% file: Chap5.tex
This paper proposes a TS-Conv method to cope with the IFS problem faced by the existing AOOD methods. The proposed TS-Conv samples the task-wise features from their respective sensitive regions and maps them together in alignment to guide a dynamic label assignment for better performance and robustness. Extensive experiments on several public datasets demonstrate the following: 1) The effectiveness of each improvement designed in TS-Conv has been confirmed, and the claims made for each component have been verified. 2) TS-Conv has good scalability on lightweight models and for multimodal data. It can improve the performance of the lightweight model without extra inference cost and be extended to sample modality-wise features with positive results. 3) The proposed TS-Conv has achieved advanced performance and speed compared to existing AOOD methods on the datasets covering multiple scenes, multimodal images, etc., demonstrating its generality further.  Nonetheless, the computational complexity of using DCN is still high, and the impact of OBB annotation error on explicit constraints of localization on TS-Conv is unclear. In the future, it would be interesting to investigate these issues.

The code is available at https://github.com/Shank2358.

%% file: SuppMate.tex

\subsection{Display of More Experiment Results}
\begin{figure}[!ht]
	\vspace{-0.5em}
	\centering
	\epsfig{width=0.48\textwidth,file=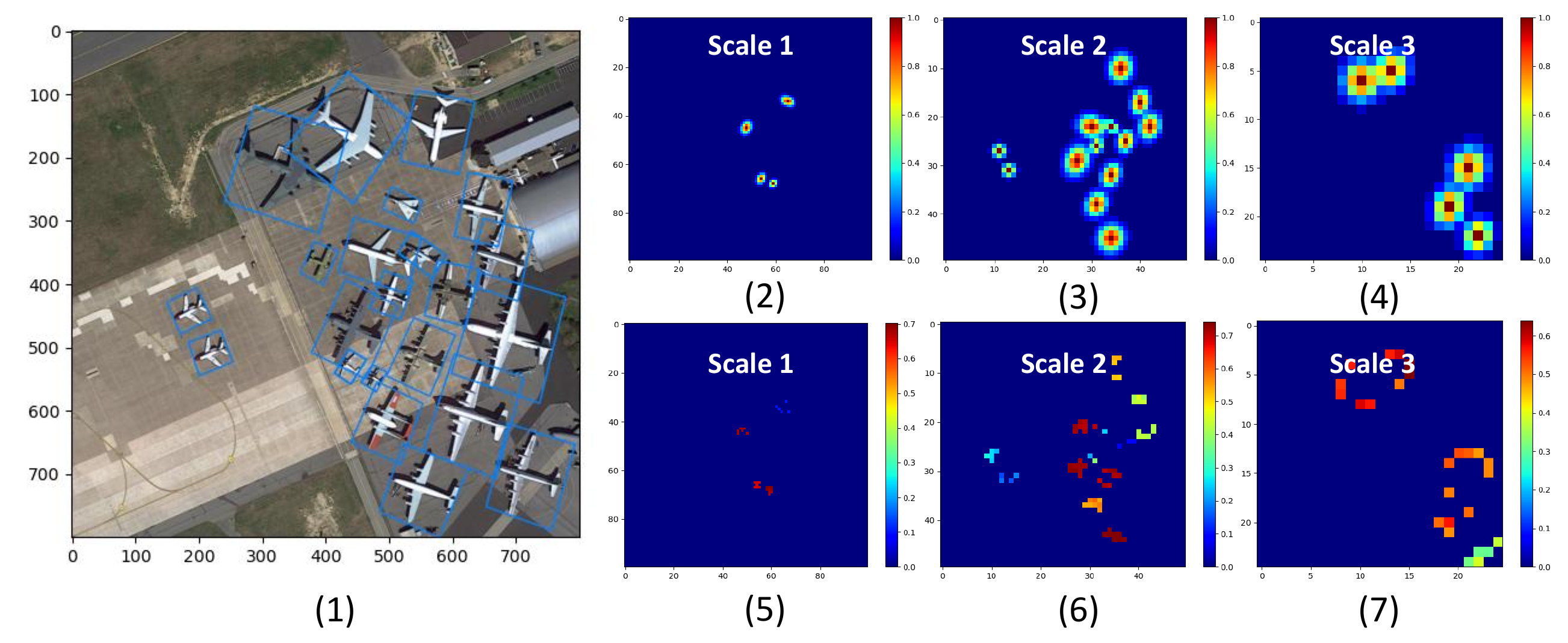}
	\vspace{-0.5em}
	\caption{Comparison of DTLA and GGHL label assignment strategies. Figs. (2)-(4) represent the positive candidate positions and their scores statically assigned by the GGHL strategy at three different scales, respectively. Figs. (5)-(7) represent the positive candidate positions and their scores dynamically assigned by the proposed DTLA strategy at three different scales, respectively. The closer the color is to red, the higher the score.}\label{fig:6RR}
\end{figure}

\begin{figure}[!ht]
	\vspace{-0.5em}
	\centering
	\epsfig{width=0.48\textwidth,file=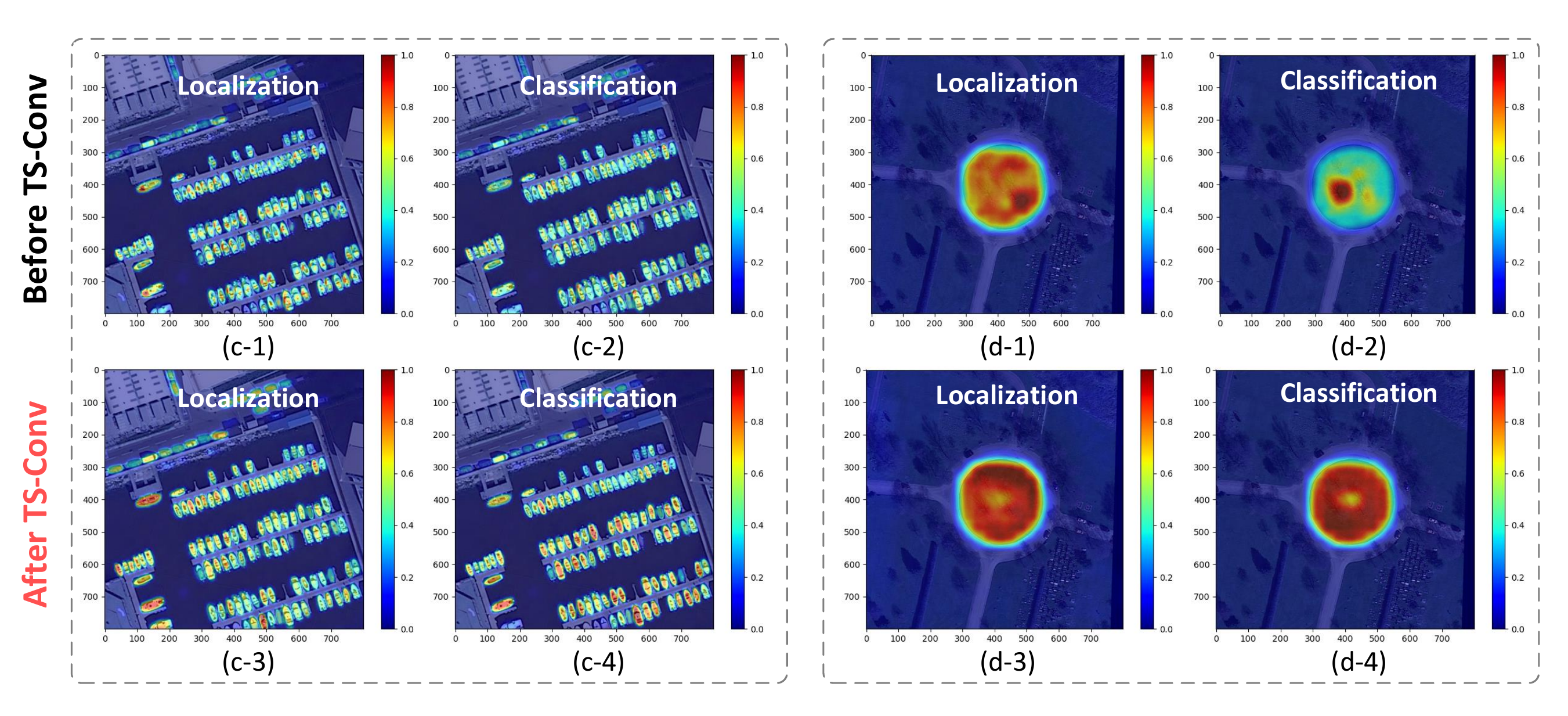}
	\vspace{-0.5em}
	\caption{Visualization of feature sensitivity regions. The closer the color is to red, the higher the sensitivity of the feature at this location. The figures in rows 1 and 3 (Figs. (c-1), (c-2), (d-1) and (d-2)) are the visualization results of feature sensitivity regions before using the designed TS-Conv. The figures in rows 2 and 4 (Figs. (c-3), (c-4), (d-3) and (d-4)) are the visualization results of feature sensitivity regions after using the designed TS-Conv.}\label{fig:7RR}
\end{figure}

\begin{figure}[!ht]
	\vspace{-0.5em}
	\centering
	\epsfig{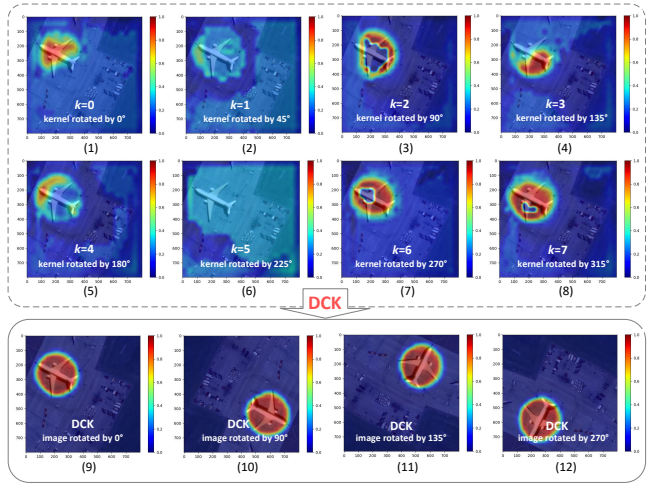}
	\vspace{-0.5em}
	\caption{Visualization of features extracted by Dynamic circular kernel (DCK). Figs. (1)-(8) show the features extracted by convolutional kernels with different rotations in DCK, respectively. Figs. (9)-(12) show the DCK-extracted features of input images with different rotations, respectively.}\label{fig:8RR}
\end{figure}

\begin{figure}[!tp]
	\centering
	\epsfig{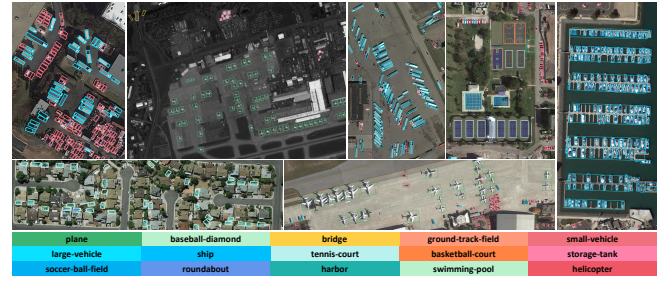}
	\caption{Visualization results of TS-Conv on the DOTA dataset.}\label{fig:11}
\end{figure}

\begin{figure}[!t]
	\centering
	\epsfig{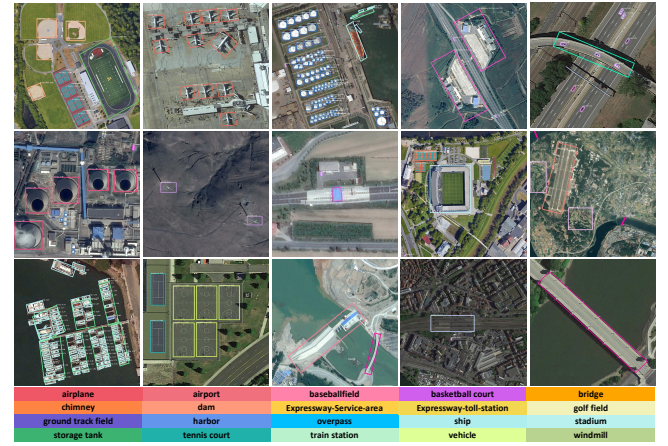}
	\caption{Visualization results of TS-Conv on the DIOR-R dataset.}\label{fig:13}
	\vspace{-0.5em}
\end{figure}

\subsection{Experiments in Other AOOD Scenes}

\begin{figure}[!ht]
	\centering
	\epsfig{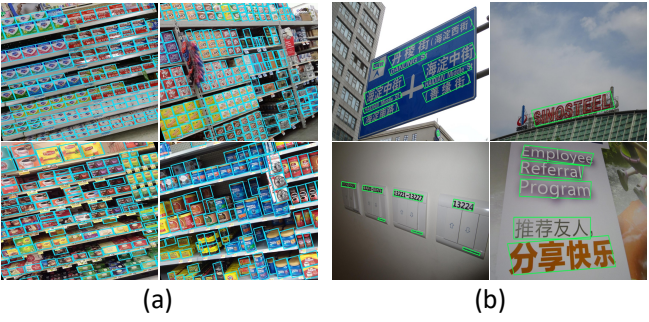}
	\caption{Visualization results of TS-Conv on a) the SKU-110KR dataset and b) the MSRA-TD500 dataset.}\label{fig:15}
\end{figure}

To further verify the generality of the proposed TS-Conv in other AOOD scenes, experiments are also conducted on the SKU-110KR \cite{pan2020dynamic} dataset and MSRA-TD500 \cite{yao2012detecting} dataset. {SKU110-R \cite{pan2020dynamic} is a dense-oriented commodity detection dataset containing 1,733,678 instances, of which the images are collected from supermarket stores. The ratio of the training set and test set in the experiment is 1:1. MSRA-TD500 \cite{yao2012detecting} is an oriented text detection dataset containing 500 images, with a training set and test set ratio of 3:2.} The results listed in Table~\ref{table:11} and Table~\ref{table:12}, respectively, show that the proposed TS-Conv also achieves better performance than existing methods in other AOOD scenes. The visualization results are shown in Fig.~\ref{fig:15}. 

\begin{table}[ht]
	\centering
	\renewcommand\arraystretch{1.3}
	\setlength{\tabcolsep}{3mm}{
		\caption{\label{table:11}
			{Comparative experiments on the SKU-110KR dataset}}
		\resizebox{0.48\textwidth}{!}{\setlength{\tabcolsep}{3mm}{
				\begin{threeparttable}
					\begin{tabular}{c|c|c|c}
						\hline\hline
						Methods & Anchor & Backbone & mAP$_{75}$ \\ \hline
						YOLOv3-R \cite{pan2020dynamic} & AB & DarkNet53 & 51.10 \\
						CenterNet-R \cite{pan2020dynamic}  & AF & Hourglass104 & 61.10 \\
						DRN \cite{pan2020dynamic}  & AF & Hourglass104 & 63.10 \\ \hline
						GGHL \cite{huang2022general} (Baseline) & AF & DarkNet53 & 63.73  \\ 
						\textbf{TS-Conv} & AF  & DarkNet53 & \textbf{65.32 \tiny{(+1.59)}} \\
						\hline\hline
					\end{tabular}
	\end{threeparttable}}}}\vspace{0.5em}
	\justifying{Note: Bold indicates the best result. AF represents anchor-free methods, and AB represents anchor-based methods.}
\end{table}
\begin{table}[!ht]
	\vspace{-0.5em}
	\centering
	\renewcommand\arraystretch{1.2}
	\setlength{\tabcolsep}{3mm}{
		\caption{\label{table:12}
			{Comparative experiments on the MSRA-TD500 dataset}}
		\resizebox{0.48\textwidth}{!}{\setlength{\tabcolsep}{3mm}{
				\begin{threeparttable}
					\begin{tabular}{c|c|c|c|c}
						\hline\hline
						Methods & mAP$_{50}$ & Precision & Recall & F-measure \\ \hline		
						EAST \cite{xu2020gliding} & - & 91.50 & 83.30 & 87.20 \\	
						Gliding Vertex \cite{xu2020gliding} & - & 88.80 & 84.30 & 86.50 \\		
						DBNet++ \cite{10042179} & - & 91.50 & 83.30 & 87.20 \\			
						CGCDet \cite{10042179} & - & 90.95 & 84.95 & 87.85 \\
						KFIoU \cite{yang2022kfiou} & 76.30 & - & - & - \\
						\hline
						GGHL \cite{huang2022general} (Baseline) & 70.41 & 89.65 & 86.21 & 87.89 \\ 
						\textbf{TS-Conv} & 74.43 \tiny{(+4.02)}  & 90.48 & 86.84 & 88.62 \\
						\textbf{TS-Conv$*$} & \textbf{77.09 \tiny{(+4.02)}} & \textbf{91.52} & \textbf{87.00} & \textbf{89.20} \\
						\hline\hline
					\end{tabular}
	\end{threeparttable}}}}\vspace{0.5em}
	\justifying{Note: Bold indicates the best result. AF represents anchor-free methods, and AB represents anchor-based methods. {“$*$” represents multi-scale training and multi-scale testing.}}
\end{table}

\subsection{Some Failure Cases}

\begin{figure}[!t]
	\centering
	\epsfig{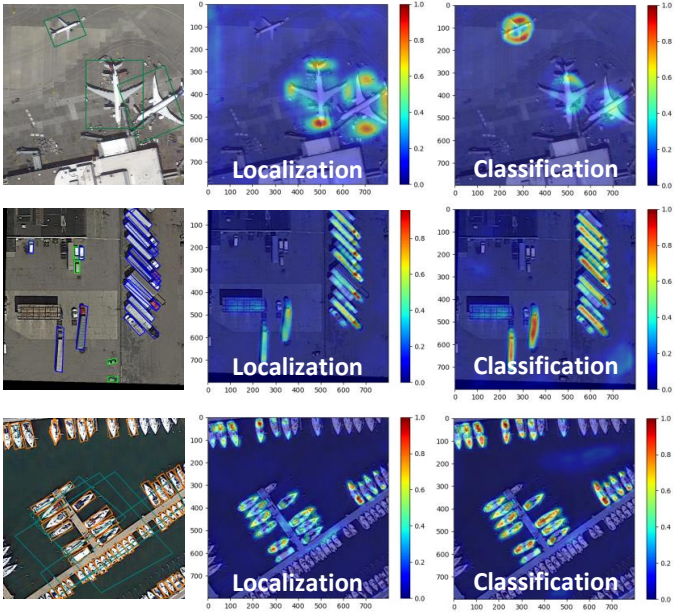}
	\vspace{-1em}
	\caption{Some failure cases of TS-Conv.}\label{fig:case}
\end{figure}

Additionally, some failure cases are provided in Fig.~\ref{fig:case} for the future research.

\subsection{Circular Kernel Generated by Bilinear Interpolation}
Define the vanilla $3 \times 3$ convolutional kernel as
\begin{equation}
	{\boldsymbol{K}^{cls}} = \left[ {\begin{array}{*{20}{c}}
			{{K_0} }&{{K_1}}&{{K_2} }\\
			{{K_3}}&{{K_4}}&{{K_5}}\\
			{{K_6} }&{{K_7}}&{{K_8} }
	\end{array}} \right].
	\tag{A-1}
	\label{eq:A-1}
\end{equation}
The $3 \times 3$ circular convolutional kernel generated by bilinear interpolation from ${\boldsymbol K}^{cls}$ is represented as
\begin{equation}
	{\boldsymbol{\dot K}^{cls}}  = \left[ {\begin{array}{*{20}{c}}
			{{\dot K_0}}&{{K_1}}&{{\dot K_2} }\\
			{{K_3}}&{{K_4}}&{{K_5}}\\
			{{\dot K_6}}&{{K_7}}&{{\dot K_8}}
	\end{array}} \right],
	\tag{A-2}
	\label{eq:A-2}
\end{equation}
where 
\begin{equation}
	\left\{ \setlength{\arraycolsep}{1pt}{\begin{array}{l}
			{\dot K_0}  = {\textstyle{1 \over 2}}{K_0} + {\textstyle{{\sqrt 2  - 1} \over 2}}{K_1} + {\textstyle{{\sqrt 2  - 1} \over 2}}{K_3} + {\textstyle{{3 - 2\sqrt 2 } \over 2}}{K_4}\\
			{\dot K_2}  = {\textstyle{{\sqrt 2  - 1} \over 2}}{K_1} + {\textstyle{1 \over 2}}{K_2} + {\textstyle{{3 - 2\sqrt 2 } \over 2}}{K_4} + {\textstyle{{\sqrt 2  - 1} \over 2}}{K_5}\\
			{\dot K_6}  = {\textstyle{{\sqrt 2  - 1} \over 2}}{K_3} + {\textstyle{{3 - 2\sqrt 2 } \over 2}}{K_4} + {\textstyle{1 \over 2}}{K_6} + {\textstyle{{\sqrt 2  - 1} \over 2}}{K_7}\\
			{\dot K_8}  = {\textstyle{{3 - 2\sqrt 2 } \over 2}}{K_4} + {\textstyle{{\sqrt 2  - 1} \over 2}}{K_5} + {\textstyle{{\sqrt 2  - 1} \over 2}}{K_7} + {\textstyle{1 \over 2}}{K_8}
	\end{array}} \right..
	\tag{A-3}
	\label{eq:A-3}
\end{equation}

\subsection{GGHL Label Assignment Strategy}
Define the Gaussian probability density function (PDF) of the candidate position $\left( {x,y} \right)$ generated by GGHL \cite{huang2022general} as
\begin{equation}
	f\left(x, y \right) = \frac{1}{{ \sqrt {2\pi \boldsymbol Q \boldsymbol \Lambda {\boldsymbol Q^T}} }} \times { e^{-\!\frac{1}{2}{		
				{{\left( {\boldsymbol X - \boldsymbol u} \right)}^T}{\boldsymbol C^{ - 1}}\left( {\boldsymbol X - \boldsymbol u} \right)
	}}},
	\tag{B-1}
	\label{eq:B-1}
\end{equation}
\begin{equation}
	\boldsymbol C = \boldsymbol A{\boldsymbol A^T} = \boldsymbol Q \boldsymbol \Lambda {\boldsymbol Q^T} = \left( {\boldsymbol Q{\boldsymbol \Lambda ^{1/2}}} \right){\left( {\boldsymbol Q{\boldsymbol \Lambda ^{1/2}}} \right)^T},
	\tag{B-2}
	\label{eq:B-2}
\end{equation}
where $\boldsymbol{X} = {\left[ {x,y} \right]^T}$. In this case, the mean vector $\boldsymbol \mu  = {\left[ {{x_c},{y_c}} \right]^T}$ controls the spatial translation, the real orthogonal matrix $\boldsymbol Q = \small {\left[ \setlength{\arraycolsep}{3pt}{{\begin{array}{*{10}{c}}
				{\cos \alpha }&{ -\sin \alpha }\\
				{\sin \alpha }&{\cos \alpha }
	\end{array}}} \right]}$ is a rotation matrix, and the diagonal matrix
$\boldsymbol \Lambda = \small {\left[ \setlength{\arraycolsep}{3pt}{{\begin{array}{*{10}{c}}
				{{{{\left( {{\textstyle{{{S_1}} \over 2}}} \right)}^2}}}&{}\\
				{}&{{{{\left( {{\textstyle{{{S_2}} \over 2}}} \right)}^2}}}
	\end{array}}} \right]}$
represents the scaling. To make the value of $f\left(x, y \right)$ in the range $\left(0, 1 \right)$, the constant coefficient term, $\xi =\! \frac{1}{{\! \sqrt {2\pi \! \boldsymbol Q \! \boldsymbol \Lambda {\boldsymbol Q^T}} }}$ of $f\left(x, y \right)$ is removed to obtain Gaussian heatmap score $F_{x,y}$.

%% file: main.bbl
\begin{thebibliography}{10}
\providecommand{\url}[1]{#1}
\csname url@rmstyle\endcsname
\providecommand{\newblock}{\relax}
\providecommand{\bibinfo}[2]{#2}
\providecommand\BIBentrySTDinterwordspacing{\spaceskip=0pt\relax}
\providecommand\BIBentryALTinterwordstretchfactor{4}
\providecommand\BIBentryALTinterwordspacing{\spaceskip=\fontdimen2\font plus
\BIBentryALTinterwordstretchfactor\fontdimen3\font minus
  \fontdimen4\font\relax}
\providecommand\BIBforeignlanguage[2]{{%
\expandafter\ifx\csname l@#1\endcsname\relax
\typeout{** WARNING: IEEEtran.bst: No hyphenation pattern has been}%
\typeout{** loaded for the language `#1'. Using the pattern for}%
\typeout{** the default language instead.}%
\else
\language=\csname l@#1\endcsname
\fi
#2}}

\bibitem{xiaDOTALargeScaleDataset2018}
G.-S. Xia, X.~Bai, J.~Ding, Z.~Zhu, S.~Belongie, J.~Luo, M.~Datcu, M.~Pelillo,
  and L.~Zhang, ``{{DOTA}}: {{A}} large-scale dataset for object detection in
  aerial images,'' in \emph{Proc. IEEE/CVF Conf. Comput. Vis. Pattern
  Recognit.}, {Salt Lake City, Utah, USA}, June 2018, pp. 3974--3983.

\bibitem{renFasterRCNNRealTime2017a}
S.~Ren, K.~He, R.~Girshick, and J.~Sun, ``Faster {{R-CNN}}: {{Towards}}
  real-time object detection with region proposal networks,'' \emph{IEEE Trans.
  Pattern Anal. Mach. Intell.}, vol.~39, no.~6, pp. 1137--1149, 2017.

\bibitem{bai2023localizing}
{J. Bai, J. Ren, Z. Xiao, Z. Chen, C. Gao, T. A. A. Ali, and L. Jiao},
  ``{Localizing from classification: self-directed weakly supervised object
  localization for remote sensing images},'' \emph{{IEEE Trans. Neural Netw.
  Learn. Syst.}}, pp. {1--15}, {2023}.

\bibitem{ding2019learning}
J.~Ding, N.~Xue, Y.~Long, G.-S. Xia, and Q.~Lu, ``Learning {{RoI}} transformer
  for oriented object detection in aerial images,'' in \emph{Proc. IEEE/CVF
  Conf. Comput. Vis. Pattern Recognit.}, {Long Beach, CA, USA}, June 2019, pp.
  2849--2858.

\bibitem{xu2020gliding}
Y.~Xu, M.~Fu, Q.~Wang, Y.~Wang, K.~Chen, G.-S. Xia, and X.~Bai, ``Gliding
  vertex on the horizontal bounding box for multi-oriented object detection,''
  \emph{IEEE Trans. Pattern Anal. Mach. Intell.}, pp. 1--1, Feb. 2020.

\bibitem{yang2019r3det}
X.~Yang, Q.~Liu, J.~Yan, and A.~Li, ``{{R3Det}}: {{Refined}} single-stage
  detector with feature refinement for rotating object,'' \emph{arXiv preprint
  arXiv:1908.05612}, 2019.

\bibitem{huang2022general}
Z.~Huang, W.~Li, X.-G. Xia, and R.~Tao, ``A general gaussian heatmap label
  assignment for arbitrary-oriented object detection,'' \emph{IEEE Trans. Image
  Process.}, vol.~31, pp. 1895--1910, 2022.

\bibitem{jiang2018acquisition}
B.~Jiang, R.~Luo, J.~Mao, T.~Xiao, and Y.~Jiang, ``Acquisition of localization
  confidence for accurate object detection,'' in \emph{Proc. Eur. Conf. Comput.
  Vis.}, 2018, pp. 784--799.

\bibitem{song2020revisiting}
G.~Song, Y.~Liu, and X.~Wang, ``Revisiting the sibling head in object
  detector,'' in \emph{Proc. IEEE/CVF Conf. Comput. Vis. Pattern Recognit.},
  2020, pp. 11\,563--11\,572.

\bibitem{piao2023unsupervised}
{Z. Piao, L. Tang, and B. Zhao}, ``{Unsupervised domain-adaptive object
  detection via localization regression alignment},'' \emph{{IEEE Trans. Neural
  Netw. Learn. Syst.}}, pp. {1--15}, {2023}.

\bibitem{han2021redet}
J.~Han, J.~Ding, N.~Xue, and G.-S. Xia, ``{{ReDet}}: {{A}} rotation-equivariant
  detector for aerial object detection,'' \emph{arXiv preprint
  arXiv:2103.07733}, 2021.

\bibitem{yangSCRDetMoreRobust2019}
X.~Yang, J.~Yang, J.~Yan, Y.~Zhang, T.~Zhang, Z.~Guo, X.~Sun, and K.~Fu,
  ``{{SCRDet}}: {{Towards}} more robust detection for small, cluttered and
  rotated objects,'' in \emph{Proc. IEEE Int. Conf. Comput. Vis.}, {Seoul,
  South Korea}, Oct. 2019, pp. 8232--8241.

\bibitem{yang2019reppoints}
Z.~Yang, S.~Liu, H.~Hu, L.~Wang, and S.~Lin, ``Reppoints: {{Point}} set
  representation for object detection,'' in \emph{Proc. IEEE Int. Conf. Comput.
  Vis.}, 2019, pp. 9657--9666.

\bibitem{li2022oriented}
W.~Li, Y.~Chen, K.~Hu, and J.~Zhu, ``Oriented reppoints for aerial object
  detection,'' in \emph{Proc. IEEE/CVF Conf. Comput. Vis. Pattern Recognit.},
  2022, pp. 1829--1838.

\bibitem{zhu2020autoassign}
B.~Zhu, J.~Wang, Z.~Jiang, F.~Zong, S.~Liu, Z.~Li, and J.~Sun, ``Autoassign:
  Differentiable label assignment for dense object detection,'' \emph{arXiv
  preprint arXiv:2007.03496}, 2020.

\bibitem{zhang2020bridging}
S.~Zhang, C.~Chi, Y.~Yao, Z.~Lei, and S.~Z. Li, ``Bridging the gap between
  anchor-based and anchor-free detection via adaptive training sample
  selection,'' in \emph{Proc. IEEE/CVF Conf. Comput. Vis. Pattern Recognit.},
  2020, pp. 9759--9768.

\bibitem{han2021align}
J.~Han, J.~Ding, J.~Li, and G.-S. Xia, ``Align deep features for oriented
  object detection,'' \emph{IEEE Trans. Geosci. Remote Sens.}, pp. 1--11, 2021.

\bibitem{yang2020arbitrary}
X.~Yang and J.~Yan, ``Arbitrary-oriented object detection with circular smooth
  label,'' in \emph{Proc. Eur. Conf. Comput. Vis.}, {Online}, Aug. 2020, pp.
  677--694.

\bibitem{yi2020oriented}
J.~Yi, P.~Wu, B.~Liu, Q.~Huang, H.~Qu, and D.~N. Metaxas, ``Oriented object
  detection in aerial images with box boundary-aware vectors,'' in \emph{Proc.
  IEEE Winter Conf. Appl. Comput. Vis.}, Dec. 2020, pp. 2150--2159.

\bibitem{wei2020oriented}
H.~Wei, Y.~Zhang, Z.~Chang, H.~Li, H.~Wang, and X.~Sun, ``Oriented objects as
  pairs of middle lines,'' \emph{ISPRS J. Photogramm. Remote Sens.}, vol. 169,
  pp. 268--279, 2020.

\bibitem{dai2022ao2}
L.~Dai, H.~Liu, H.~Tang, Z.~Wu, and P.~Song, ``{{AO2-DETR}}:
  {{Arbitrary-oriented}} object detection transformer,'' \emph{arXiv preprint
  arXiv:2205.12785}, 2022.

\bibitem{yang2021rethinking}
X.~Yang, J.~Yan, Q.~Ming, W.~Wang, X.~Zhang, and Q.~Tian, ``Rethinking rotated
  object detection with gaussian wasserstein distance loss,'' \emph{arXiv
  preprint arXiv:2101.11952}, 2021.

\bibitem{yang2021learning}
X.~Yang, X.~Yang, J.~Yang, Q.~Ming, W.~Wang, Q.~Tian, and J.~Yan, ``Learning
  high-precision bounding box for rotated object detection via kullback-leibler
  divergence,'' \emph{arXiv preprint arXiv:2106.01883}, 2021.

\bibitem{9578090}
Z.~Guo, C.~Liu, X.~Zhang, J.~Jiao, X.~Ji, and Q.~Ye, ``Beyond bounding-box:
  Convex-hull feature adaptation for oriented and densely packed object
  detection,'' in \emph{Proc. IEEE/CVF Conf. Comput. Vis. Pattern Recognit.},
  2021, pp. 8788--8797.

\bibitem{Xu_2023_CVPR}
C.~Xu, J.~Ding, J.~Wang, W.~Yang, H.~Yu, L.~Yu, and G.-S. Xia, ``Dynamic
  coarse-to-fine learning for oriented tiny object detection,'' in \emph{Proc.
  IEEE/CVF Conf. Comput. Vis. Pattern Recognit.}, June 2023, pp. 7318--7328.

\bibitem{ming2021cfc}
Q.~Ming, L.~Miao, Z.~Zhou, and Y.~Dong, ``{{CFC-Net}}: {{A}} critical feature
  capturing network for arbitrary-oriented object detection in remote sensing
  images,'' \emph{arXiv preprint arXiv:2101.06849}, 2021.

\bibitem{zhang2021varifocalnet}
H.~Zhang, Y.~Wang, F.~Dayoub, and N.~Sunderhauf, ``Varifocalnet: {{An}}
  iou-aware dense object detector,'' in \emph{Proc. IEEE/CVF Conf. Comput. Vis.
  Pattern Recognit.}, 2021, pp. 8514--8523.

\bibitem{wu2020rethinking}
Y.~Wu, Y.~Chen, L.~Yuan, Z.~Liu, L.~Wang, H.~Li, and Y.~Fu, ``Rethinking
  classification and localization for object detection,'' in \emph{Proc.
  IEEE/CVF Conf. Comput. Vis. Pattern Recognit.}, 2020, pp. 10\,186--10\,195.

\bibitem{ge2021yolox}
Z.~Ge, S.~Liu, F.~Wang, Z.~Li, and J.~Sun, ``Yolox: {{Exceeding}} yolo series
  in 2021,'' \emph{arXiv preprint arXiv:2107.08430}, 2021.

\bibitem{cao2020d2det}
J.~Cao, H.~Cholakkal, R.~M. Anwer, F.~S. Khan, Y.~Pang, and L.~Shao, ``D2det:
  {{Towards}} high quality object detection and instance segmentation,'' in
  \emph{Proc. IEEE/CVF Conf. Comput. Vis. Pattern Recognit.}, 2020, pp.
  11\,485--11\,494.

\bibitem{zhu2019deformable}
X.~Zhu, H.~Hu, S.~Lin, and J.~Dai, ``Deformable convnets v2: {{More}}
  deformable, better results,'' in \emph{Proc. IEEE/CVF Conf. Comput. Vis.
  Pattern Recognit.}, 2019, pp. 9308--9316.

\bibitem{wang2019region}
J.~Wang, K.~Chen, S.~Yang, C.~C. Loy, and D.~Lin, ``Region proposal by guided
  anchoring,'' in \emph{Proc. IEEE/CVF Conf. Comput. Vis. Pattern Recognit.},
  2019, pp. 2965--2974.

\bibitem{dai2017deformable}
J.~Dai, H.~Qi, Y.~Xiong, Y.~Li, G.~Zhang, H.~Hu, and Y.~Wei, ``Deformable
  convolutional networks,'' in \emph{Proc. IEEE Int. Conf. Comput. Vis.}, 2017,
  pp. 764--773.

\bibitem{pan2020dynamic}
X.~Pan, Y.~Ren, K.~Sheng, W.~Dong, H.~Yuan, X.~Guo, C.~Ma, and C.~Xu, ``Dynamic
  refinement network for oriented and densely packed object detection,'' in
  \emph{Proc. IEEE/CVF Conf. Comput. Vis. Pattern Recognit.}, {Online}, June
  2020, pp. 11\,207--11\,216.

\bibitem{liu2018intriguing}
R.~Liu, J.~Lehman, P.~Molino, F.~Petroski~Such, E.~Frank, A.~Sergeev, and
  J.~Yosinski, ``An intriguing failing of convolutional neural networks and the
  coordconv solution,'' \emph{Adv. Neur. In.}, vol.~31, 2018.

\bibitem{9669124}
{S.-C. Huang, Q.-V. Hoang, and T.-H. Le}, ``{SFA-Net: A selective features
  absorption network for object detection in rainy weather conditions},''
  \emph{{IEEE Trans. Neural Netw. Learn. Syst.}}, vol.~{34}, no.~{8}, pp.
  {5122--5132}, {2023}.

\bibitem{9733170}
{Z. Shao, J. Han, and D. Marnerides, and K. Debattista}, ``{Region-object
  relation-aware dense captioning via transformer},'' \emph{{IEEE Trans. Neural
  Netw. Learn. Syst.}}, pp. {1--12}, {2022}.

\bibitem{yang2019condconv}
B.~Yang, G.~Bender, Q.~V. Le, and J.~Ngiam, ``Condconv: {{Conditionally}}
  parameterized convolutions for efficient inference,'' \emph{Adv. Neur. In.},
  vol.~32, 2019.

\bibitem{rezatofighiGeneralizedIntersectionUnion2019}
H.~Rezatofighi, N.~Tsoi, J.~Gwak, A.~Sadeghian, I.~Reid, and S.~Savarese,
  ``Generalized intersection over union: {{A}} metric and a loss for bounding
  box regression,'' in \emph{Proc. IEEE/CVF Conf. Comput. Vis. Pattern
  Recognit.}, {Long Beach, CA, USA}, June 2019, pp. 658--666.

\bibitem{linFocalLossDense2017}
T.-Y. Lin, P.~Goyal, R.~Girshick, K.~He, and P.~Dollar, ``Focal loss for dense
  object detection,'' in \emph{Proc. IEEE Int. Conf. Comput. Vis.}, {Venice,
  Italy}, Oct. 2017, pp. 2999--3007.

\bibitem{huang2021lo}
Z.~Huang, W.~Li, X.-G. Xia, H.~Wang, F.~Jie, and R.~Tao, ``{{LO-Det}}:
  {{Lightweight}} oriented object detection in remote sensing images,''
  \emph{IEEE Trans. Geosci. Remote Sens.}, pp. 1--15, 2021.

\bibitem{li2020object}
K.~Li, G.~Wan, G.~Cheng, L.~Meng, and J.~Han, ``Object detection in optical
  remote sensing images: {{A}} survey and a new benchmark,'' \emph{ISPRS J.
  Photogramm. Remote Sens.}, vol. 159, pp. 296--307, 2020.

\bibitem{liu2017high}
Z.~Liu, L.~Yuan, L.~Weng, and Y.~Yang, ``A high resolution optical satellite
  image dataset for ship recognition and some new baselines,'' in \emph{Proc.
  Int. Conf. Pattern Recognit. Appl. Methods}, vol.~2.\hskip 1em plus 0.5em
  minus 0.4em\relax {SciTePress}, 2017, pp. 324--331.

\bibitem{ding2021object}
J.~Ding, N.~Xue, G.-S. Xia, X.~Bai, W.~Yang, M.~Y. Yang, S.~Belongie, J.~Luo,
  M.~Datcu, M.~Pelillo, \emph{et~al.}, ``Object detection in aerial images:
  {{A}} large-scale benchmark and challenges,'' \emph{arXiv preprint
  arXiv:2102.12219}, 2021.

\bibitem{cheng2022anchor}
G.~Cheng, J.~Wang, K.~Li, X.~Xie, C.~Lang, Y.~Yao, and J.~Han, ``Anchor-free
  oriented proposal generator for object detection,'' \emph{IEEE Trans. Geosci.
  Remote Sens.}, vol.~60, pp. 1--11, 2022.

\bibitem{sun2022drone}
Y.~Sun, B.~Cao, P.~Zhu, and Q.~Hu, ``Drone-based {{RGB-Infrared}}
  cross-modality vehicle detection via uncertainty-aware learning,'' \emph{IEEE
  Trans. Circuits Syst. Video Technol.}, 2022.

\bibitem{li2017ship}
J.~Li, C.~Qu, and J.~Shao, ``Ship detection in sar images based on an improved
  faster r-cnn,'' in \emph{Proc. SAR Big Data Era: Model., Methods Appl.},
  {Beijing, China}, Nov. 2017, pp. 1--6.

\bibitem{9710724}
C.~Feng, Y.~Zhong, Y.~Gao, M.~R. Scott, and W.~Huang, ``Tood: Task-aligned
  one-stage object detection,'' in \emph{Proc. IEEE Int. Conf. Comput. Vis.},
  2021, pp. 3490--3499.

\bibitem{redmonYOLOv3IncrementalImprovement2018}
J.~Redmon and A.~Farhadi, ``{{YOLOv3}}: {{An}} incremental improvement,''
  \emph{arXiv preprint arXiv:1804.02767}, 2018.

\bibitem{tian2019fcos}
Z.~Tian, C.~Shen, H.~Chen, and T.~He, ``{{FCOS}}: {{Fully}} convolutional
  one-stage object detection,'' in \emph{Proc. IEEE Int. Conf. Comput. Vis.},
  {Seoul, South Korea}, Oct. 2019, pp. 9627--9636.

\bibitem{huang2022extracting}
Z.~Huang, W.~Li, and R.~Tao, ``Extracting and distilling direction-adaptive
  knowledge for lightweight object detection in remote sensing images,'' in
  \emph{Proc. IEEE Int. Conf. Acoust., Speech, Signal Process.}\hskip 1em plus
  0.5em minus 0.4em\relax {IEEE}, 2022, pp. 2360--2364.

\bibitem{qian2019learning}
W.~Qian, X.~Yang, S.~Peng, Y.~Guo, and C.~Yan, ``Learning modulated loss for
  rotated object detection,'' \emph{arXiv preprint arXiv:1911.08299}, 2019.

\bibitem{xie2021oriented}
X.~Xie, G.~Cheng, J.~Wang, X.~Yao, and J.~Han, ``Oriented {{R-CNN}} for object
  detection,'' in \emph{Proc. IEEE Int. Conf. Comput. Vis.}, 2021, pp.
  3520--3529.

\bibitem{10042179}
{Y. Wang, and Z. Zhang, W. Xu, L. Chen, G. Wang, L. Yan, S. Zhong, and X. Zou},
  ``{Learning oriented object detection via naive geometric computing},''
  \emph{{IEEE Trans. Neural Netw. Learn. Syst.}}, pp. {1--13}, {2023}.

\bibitem{9706434}
G.~Cheng, Y.~Yao, S.~Li, K.~Li, X.~Xie, J.~Wang, X.~Yao, and J.~Han,
  ``Dual-aligned oriented detector,'' \emph{IEEE Trans. Geosci. Remote Sens.},
  vol.~60, pp. 1--11, 2022.

\bibitem{ming2021optimization}
Q.~Ming, Z.~Zhou, L.~Miao, X.~Yang, and Y.~Dong, ``Optimization for oriented
  object detection via representation invariance loss,'' \emph{arXiv preprint
  arXiv:2103.11636}, 2021.

\bibitem{zhao2021polardet}
P.~Zhao, Z.~Qu, Y.~Bu, W.~Tan, and Q.~Guan, ``Polardet: {{A}} fast, more
  precise detector for rotated target in aerial images,'' \emph{Int. J. Remote
  Sens.}, vol.~42, no.~15, pp. 5831--5861, 2021.

\bibitem{yang2022kfiou}
X.~Yang, Y.~Zhou, G.~Zhang, J.~Yang, W.~Wang, J.~Yan, X.~Zhang, and Q.~Tian,
  ``The {{KFIoU}} loss for rotated object detection,'' \emph{arXiv preprint
  arXiv:2201.12558}, 2022.

\bibitem{hou2022g}
L.~Hou, K.~Lu, X.~Yang, Y.~Li, and J.~Xue, ``G-rep: {{Gaussian}} representation
  for arbitrary-oriented object detection,'' \emph{arXiv preprint
  arXiv:2205.11796}, 2022.

\bibitem{jiang2017r2cnn}
Y.~Jiang, X.~Zhu, X.~Wang, S.~Yang, W.~Li, H.~Wang, P.~Fu, and Z.~Luo,
  ``{{R2CNN}}: {{Rotational}} region {{CNN}} for orientation robust dcene text
  detection,'' \emph{arXiv preprint arXiv:1706.09579}, 2017.

\bibitem{wang2020learning}
J.~Wang, W.~Yang, H.-C. Li, H.~Zhang, and G.-S. Xia, ``Learning center
  probability map for detecting objects in aerial images,'' \emph{IEEE Trans.
  Geosci. Remote Sens.}, vol.~59, no.~5, pp. 4307--4323, 2020.

\bibitem{yang2021dense}
X.~Yang, L.~Hou, Y.~Zhou, W.~Wang, and J.~Yan, ``Dense label encoding for
  boundary discontinuity free rotation detection,'' in \emph{Proc. IEEE/CVF
  Conf. Comput. Vis. Pattern Recognit.}, 2021, pp. 15\,819--15\,829.

\bibitem{wang2018simultaneous}
J.~Wang, C.~Lu, and W.~Jiang, ``Simultaneous ship detection and orientation
  estimation in {{SAR}} images based on attention module and angle
  regression,'' \emph{Sensors-basel.}, vol.~18, no. 9-2851, pp. 1--17, 2018.

\bibitem{an2019drbox}
Q.~An, Z.~Pan, L.~Liu, and H.~You, ``{{DRBox-v2}}: {{An}} improved detector
  with rotatable boxes for target detection in {{SAR}} images,'' \emph{IEEE
  Trans. Geosci. Remote Sens.}, vol.~57, no.~11, pp. 8333--8349, 2019.

\bibitem{yao2012detecting}
C.~Yao, X.~Bai, W.~Liu, Y.~Ma, and Z.~Tu, ``Detecting texts of arbitrary
  orientations in natural images,'' in \emph{Proc. IEEE/CVF Conf. Comput. Vis.
  Pattern Recognit.}\hskip 1em plus 0.5em minus 0.4em\relax {IEEE}, 2012, pp.
  1083--1090.

\end{thebibliography}
